%% file: main.tex
\documentclass[runningheads]{llncs}

\usepackage[mobile]{eccv}

\usepackage{eccvabbrv}

\usepackage{graphicx}
\usepackage{booktabs}

\usepackage[accsupp]{axessibility}  %

\input{preamble}

\usepackage[pagebackref,breaklinks,colorlinks,citecolor=eccvblue]{hyperref}

\usepackage{orcidlink}

\begin{document}

\title{Open-Vocabulary Camouflaged Object Segmentation}

\titlerunning{OVCOS}

\author{%
  Youwei Pang\inst{1,2}\protect\footnotemark[1] \and
  Xiaoqi Zhao\inst{1,2}\protect\footnotemark[1], \\
  Jiaming Zuo\inst{2} \and
  Lihe Zhang\inst{1}\protect\footnotemark[2] \and
  Huchuan Lu\inst{1}
}

\authorrunning{Y.~Pang et al.}

\institute{%
  \small
  Dalian University of Technology \and
  X3000 Inspection Co., Ltd\\
  \texttt{\{lartpang,zxq\}@mail.dlut.edu.cn}\\
  \texttt{klaus@3000gy.com, \{zhanglihe,lhchuan\}@dlut.edu.cn}
}

\maketitle

\begin{authorfoot}
  \footnotetext[1]{Equal Contribution.}
  \footnotetext[2]{Corresponding Author.}
\end{authorfoot}

\begin{abstract}
  Recently, the emergence of the large-scale vision-language model (VLM), such as CLIP, has opened the way towards open-world object perception.
  Many works have explored the utilization of pre-trained VLM for the challenging open-vocabulary dense prediction task that requires perceiving diverse objects with novel classes at inference time.
  Existing methods construct experiments based on the public datasets of related tasks, which are not tailored for open vocabulary and rarely involve imperceptible objects camouflaged in complex scenes due to data collection bias and annotation costs.
  To fill in the gaps, we introduce a new task, open-vocabulary camouflaged object segmentation (OVCOS), and construct a large-scale complex scene dataset (\textbf{OVCamo}) containing 11,483 hand-selected images with fine annotations and corresponding object classes.
  Further, we build a strong single-stage open-vocabulary \underline{c}amouflaged \underline{o}bject \underline{s}egmentation transform\underline{er} baseline \textbf{OVCoser} attached to the parameter-fixed CLIP with iterative semantic guidance and structure enhancement.
  By integrating the guidance of class semantic knowledge and the supplement of visual structure cues from the edge and depth information, the proposed method can efficiently capture camouflaged objects.
  Moreover, this effective framework also surpasses previous state-of-the-arts of open-vocabulary semantic image segmentation by a large margin on our OVCamo dataset.
  With the proposed dataset and baseline, we hope that this new task with more practical value can further expand the research on open-vocabulary dense prediction tasks.
  Our code and data can be found in the \href{https://github.com/lartpang/OVCamo}{link}.
  \keywords{Open Vocabulary \and Camouflaged Object Segmentation}
\end{abstract}

\section{Introduction}
\label{sec:introduction}

The ability to identify and reason about object regions in a visual scene is essential for a wide range of human activities.
As a complex and fundamental task in computer vision, detecting and segmenting objects with diverse appearances in complex scenes is also an important challenge, %
which is crucial for applications across vision and robotics fields, including autonomous driving~\cite{nuScenes}, medical image analysis~\cite{Survey-MedicalImageSegmentation}, and intelligent robotics~\cite{Survey-EmbodiedVisualNavigation}, to name a few.
In the past few years, numerous typical methods~\cite{MaskRCNN,MaskFormer,Mask2Former,K-net} have emerged with the help of a mass of labeled data, which have greatly promoted the development of related fields such as semantic image segmentation (SIS)~\cite{Survey-SemanticImageSegmentation,Survey-SemanticImageSegmentationViT}.
However, existing works have mainly focused on predefined closed-set scenarios, %
where all semantic concepts are seen during both the inference and training phases.
Such a scenario setting oversimplifies the real-world complexity.
To this end, many explorations have been contributed to open vocabulary settings~\cite{Survey-OpenVocabulary,Survey-OpenVocabular2}.
In existing work, open vocabulary learning utilizes visual-related language data (e.g., object class text or descriptions) to align images and language features, enabling models to perceive novel classes without extensive labeled data~\cite{Survey-OpenVocabular2}.
Recently, large-scale pre-trained visual language models (VLMs) such as CLIP~\cite{CLIP,OpenCLIP} have been gaining attention.
Image-text matching-based learning mechanism provides them with the ability to align textual and visual signals well, and many works demonstrate their potential for open vocabulary tasks.
Besides, data plays an important role in open-vocabulary tasks.
Existing open-vocabulary semantic image segmentation (OVSIS) tasks rely on related public datasets~\cite{COCO-Stuff,ADE20K,PascalContext,MapillaryVistas,PASCAL-VOC}, while they are not designed for the open-vocabulary setting, and there is high semantic similarity between their class definitions as revealed in~\cite{OVSeg-SAN}.
Due to the data collection bias and annotation cost constraints, the existing open-vocabulary benchmarks lack special attention to finely perceive the objects of interest in concealed scenes.
And the widely used VLMs are pre-trained on image-text pairs with inherent object concept bias, thus, their ability to segment objects in complex scenes remains to be verified.

\begin{wrapfigure}{r}{0.5\linewidth}
  \vspace{-2em}
  \includegraphics[width=\linewidth]{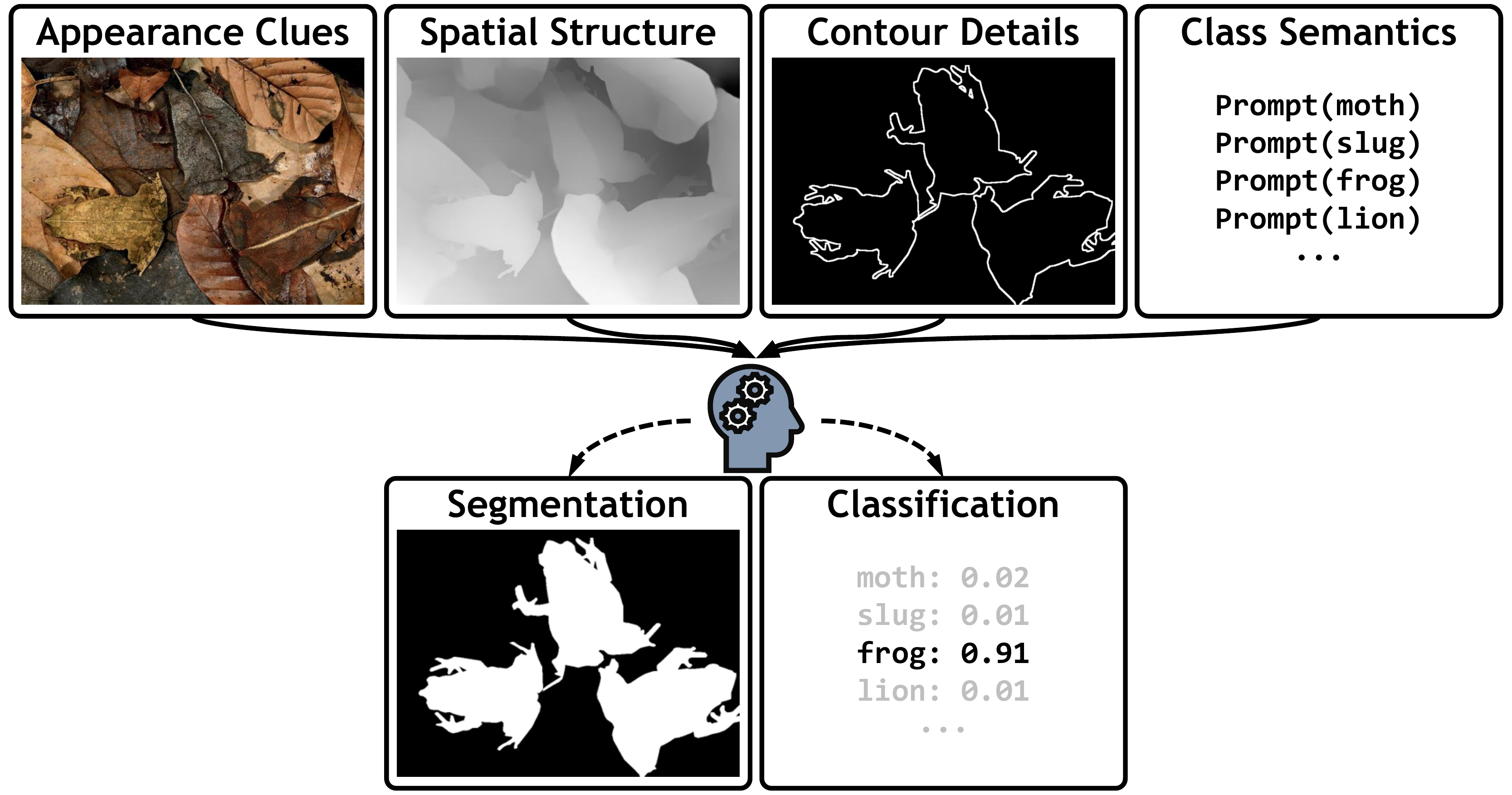}
  \caption{The perception and recognition of camouflaged objects require the collaboration of information from multiple sources such as appearance clues, spatial structure, contour details, and object semantics.}
  \label{fig:teaser2}
  \vspace{-2em}
\end{wrapfigure}
In this paper, we introduce a new open-vocabulary segmentation task \mytask dedicated to analyzing camouflaged object perception in diverse natural scenes.
And a large-scale data benchmark, named \mydataset, is carefully constructed.
Besides, we also design a strong baseline \mymethod for the proposed \mytask, based on the VLM-driven single-stage paradigm.
The camouflage\footnote{To ensure the reliability of the definition, we follow existing published literature and collect data from the \textit{class-agnostic} camouflage perception field, \ie, CSU~\cite{DeepCSU}.}
arises from several sources, including similar patterns to the environment (\eg, color and texture) and imperceptible attributes (\eg, small size and heavy occlusion) as statistically illustrated in~\cref{fig:obj-attr}.
Considering the imperceptible appearance of camouflaged objects, accurate recognition and capture actually depend more on the cooperation of multi-source knowledge.
As shown in~\cref{fig:teaser2}, in addition to visual appearance cues, we introduce the depth for the spatial structure of the scene, the edge for the regional changes about objects, and the text for the context-aware class semantics.
Considering the cooperative relationship between class recognition and object perception, the iterative learning strategy is introduced to feed back the optimized semantic relationship,
resulting in more accurate object semantic guidance.
This top-down conceptual reinforcement can further optimize open-vocabulary segmentation performance.
With the help of the iterative multi-source information joint learning strategy, our method \mymethod shows good performance in the proposed \mytask task.

In summary, our contributions are three-fold as follows:
\begin{itemize}
  \item \textbf{New Challenge.} In view of the limitations of the existing OVSIS, we introduce a more challenging \mytask task for open-vocabulary segmentation of camouflaged objects. %
  \item \textbf{New Benchmark.} A new large-scale benchmark \mydataset with diverse samples carefully collected from existing publicly available data is proposed to better evaluate and analyze the generalization of algorithms on the proposed task.%
  \item \textbf{Strong Baseline.} We build a robust single-stage baseline based on CLIP, in which the proposed iterative \mynetcore are embedded. Under the joint optimization of multi-source information, our approach \mymethod outperforms existing OVSIS algorithms on the new benchmark.
\end{itemize}

\begin{figure}[t]
  \centering
  \includegraphics[width=\linewidth]{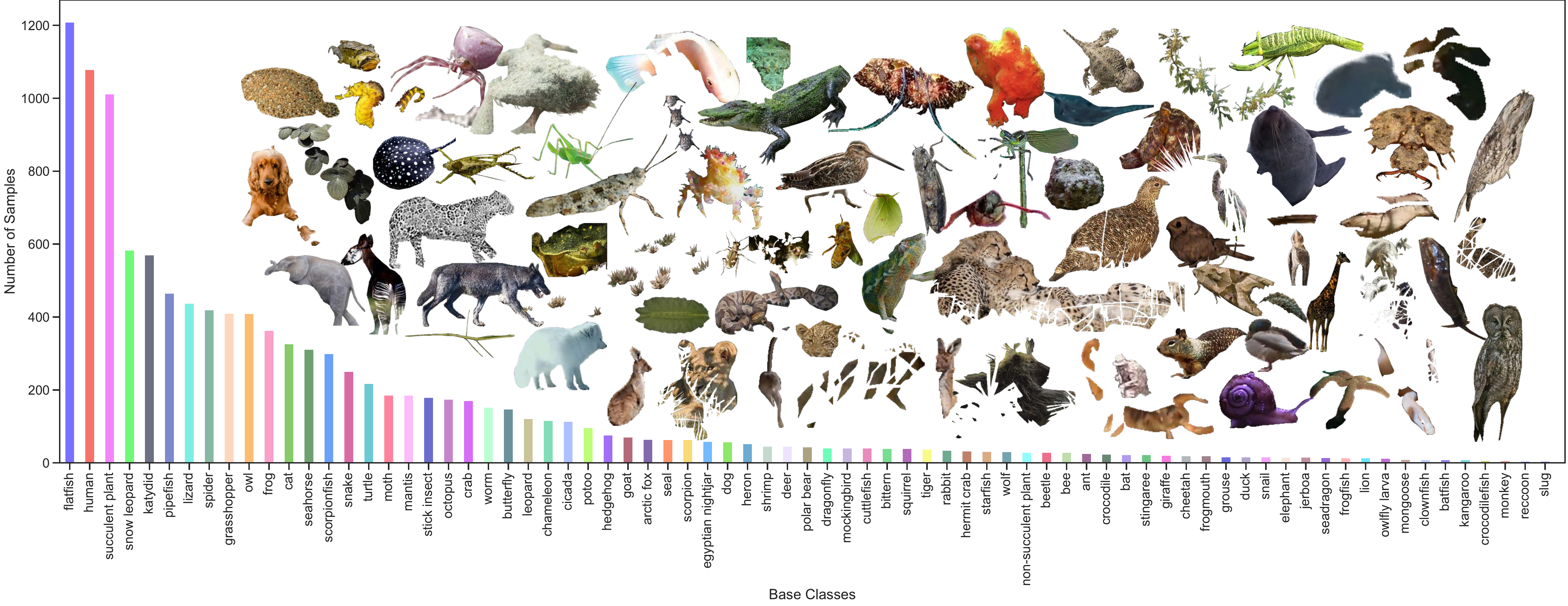}
  \caption{Class distribution in the proposed \mydataset dataset and some visual examples.}
  \label{fig:num_classes}
\end{figure}

\section{Related Works}
\label{sec:relatedworks}

\parhead{Vision-Language Pre-training.}
The core goal of visual-language pre-training is to learn generic representations of vision and language, and connect visual and language concepts.
Early approaches~\cite{VLM-Oscar,VLM-Unicoder-VL,VLM-Uniter,VL-BERT} are based on some relatively small and clean public datasets, which limits their achievable performance, and their fine-tuning on specific downstream tasks hinders application flexibility.
In~\cite{WeblySupervision}, image-text pairs collected from the Internet bring clear performance improvement for the retrieval task.
This also indirectly encourages subsequent further exploration, such as CLIP~\cite{CLIP,OpenCLIP}.
It benefits from the larger scale of noisy data from web pages covering diverse and rich concepts and exhibits impressive open-vocabulary capabilities.
This work introduces CLIP as the bedrock of open-vocabulary capability and borrows its strong image-text matching capability to build an effective baseline for the \mytask task.

\parhead{Open Vocabulary Semantic Image Segmentation (OVSIS).}
Although a variety of pipelines have emerged as summarized in~\cite{Survey-OpenVocabulary}, from an overall perspective, their efforts are similar, namely, \textit{how to align class name/description semantic embedding with visual features to anchor relevant object cues in the representation space}.
The early pioneer work~\cite{OVSeg-OVSP} attempts to connect word concepts and semantic relations, and encodes word concept hierarchy to parse images.
Due to the leading performance of the VLM on the image-text joint modeling, it has been gradually applied in the OVSIS field.
In terms of structure, existing schemes can be roughly categorized into two types:
two-stage~\cite{OVSeg-PseudoLabel,OVSeg-SimSeg,OVSeg-OVSeg,OVSeg-ODISE}
and single-stage~\cite{OVSeg-MaskCLIP,OVSeg-CATSeg,OVSeg-FCCLIP}.
In~\cite{OVSeg-PseudoLabel}, a rough segmentation map for each class is created from a VLM and then refined by the test-time augmentation. These rough maps are utilized as pseudo-labels for subsequent fine segmentation by stochastic pixel sampling.
SimSeg~\cite{OVSeg-SimSeg} adopts a cascaded design including class-agnostic proposal generation by MaskFormer~\cite{MaskFormer} and class assignment by CLIP~\cite{CLIP}.
Furthermore, OVSeg~\cite{OVSeg-OVSeg} fine-tunes CLIP on the noisy but diverse data to improve its generalization to masked images.
In lieu of using existing SIS models, a text-to-image diffusion model is introduced to generate mask features with implicit image captions in~\cite{OVSeg-ODISE}.
The single-stage design is more flexible and simpler.
MaskCLIP~\cite{OVSeg-MaskCLIP} directly modifies CLIP for semantic segmentation without training, while SAN~\cite{OVSeg-SAN} achieves better performance with the help of adapters.
CAT-Seg~\cite{OVSeg-CATSeg} highlights the importance of the cost aggregation between image and text embeddings for the OVSIS decoding.
Recent FC-CLIP~\cite{OVSeg-FCCLIP} investigates the hierarchical CLIP image encoder.
Although these methods show impressive OVSIS performance, they still follow the generalized segmentation paradigm~\cite{MaskFormer,Mask2Former} in SIS, ignoring valuable auxiliary cues for object perception.
This also causes them to struggle with objects camouflaged in complex scenes.
Unlike them, our method, which is tailored for \mytask, can effectively tap into the camouflaged object by integrating task-specific multi-source knowledge from visual appearance, spatial structure, object contour, and class semantics.

\parhead{Camouflaged Scene Understanding (CSU).}
This is a research hotspot in the computer vision community, aiming to perceive objects with camouflage~\cite{DeepCSU}.
Different from traditional object perception like salient object detection~\cite{RGBDSOD_CoNet,VSOD_DCFNet,RGBDVSOD_DVSOD,LFSOD_LFNet,DMRA_TIP,RGBDSOD_C2DFNet,RGBDSOD_DCBF,RGBSOD-MINet,RGBDSOD-HDFNet,RGBDSOD-DANet,RGBSOD-GateNet,RGBDSOD-SSLSOD,DIS-MVANet,CDCU-Spider,RGBDSOD-MMFT,RGBDTSOD-CAVER,BiSeg-GateNetv2} and semantic object segmentation~\cite{MultiSpectralVSS,ZSVOS-MSAPS-jnl}, CSU is obviously a more challenging problem.
It can be applied in some specific fields, such as medical analysis~\cite{SINetV2,Inf-Net} and agricultural management~\cite{CODRelatedWork-FruitRipenessClassification,CODRelatedWork-PestDetection}.
This topic is currently defined as the class-agnostic form, focusing on the area of camouflaged objects in the visual scene.
The available work~\cite{COD10K,UJSC,COD-SegMaR,COD-ZoomNet,COD-ZoomNeXt,SINetV2,VCOD-MoCA-Mask,COD-CamoFormer} to date has demonstrated promising performance on existing data benchmarks~\cite{CHAMELEON,CAMO,COD10K,SLSR,VCOD-CAD,VCOD-MoCA-Mask}.
Unlike previous settings, the proposed \mytask task requires further perception of object classes.
Admittedly, the publicly available data provides critical support for this new task.
It helps us take the first step.

\section{\mydataset Dataset}
\label{sec:dataset}

This work focuses on open-vocabulary segmentation in the camouflaged scene.
It enriches the connotation of OVSIS and provides a more challenging benchmark.
Note that our aim is not to claim novelty through class annotations alone, but rather to focus on how such annotations can pave the way for new tasks and methods.
Our dataset provides additional features such as clean data, less ambiguity, and sensible class hierarchies, which provide the necessary data for solving the new and complex task.
It not only adds complexity and challenge, but also promotes innovation and advances the field through the need for advanced methods.
We believe that this dataset can lead to new opportunities in areas such as wildlife conservation and intelligence sensing.

\begin{figure}[t]
  \centering
  \begin{subfigure}{0.64\linewidth}
    \centering
    \begin{subfigure}{0.49\linewidth}
      \centering
      \includegraphics[width=\linewidth]{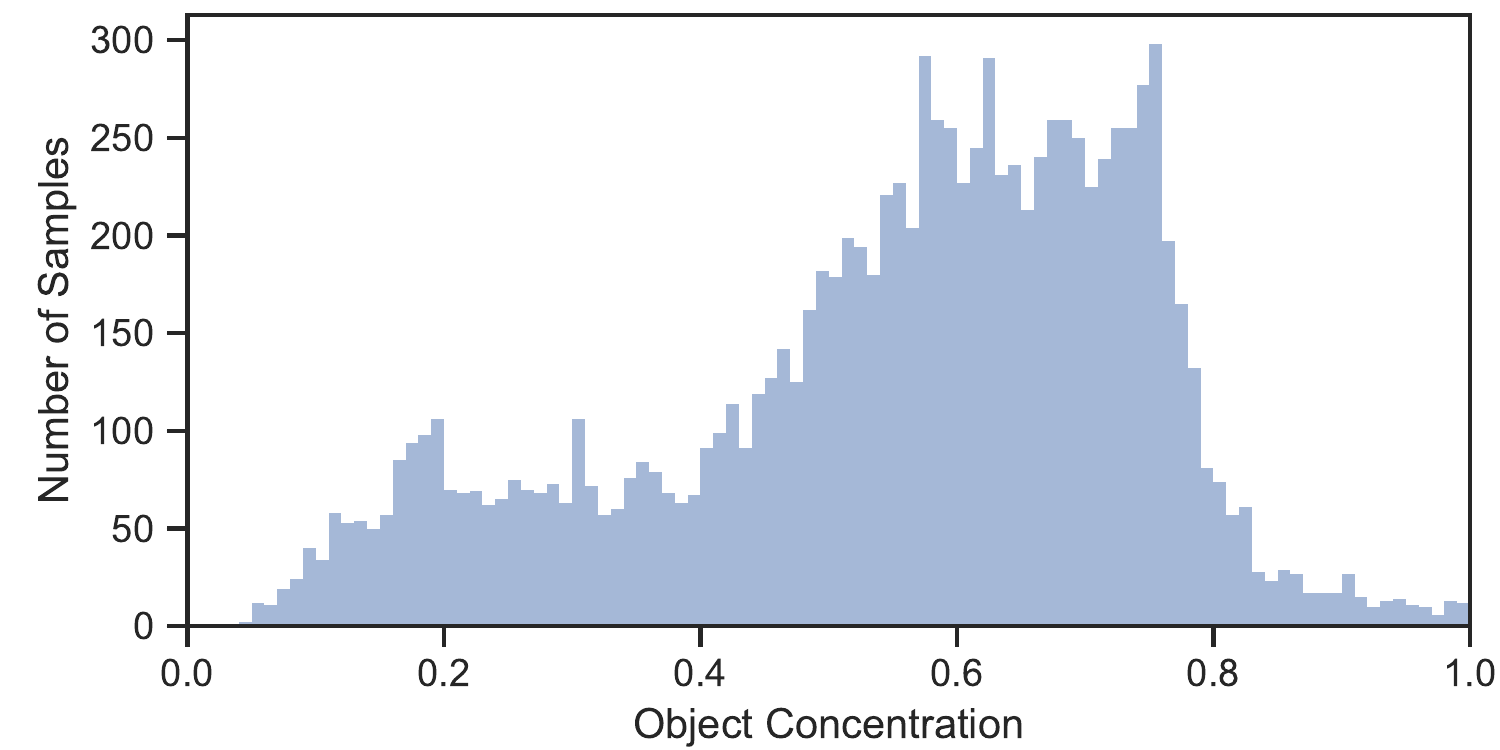}
      \caption{Object Concentration.}
      \label{fig:object_concertration}
    \end{subfigure}
    \begin{subfigure}{0.49\linewidth}
      \centering
      \includegraphics[width=\linewidth]{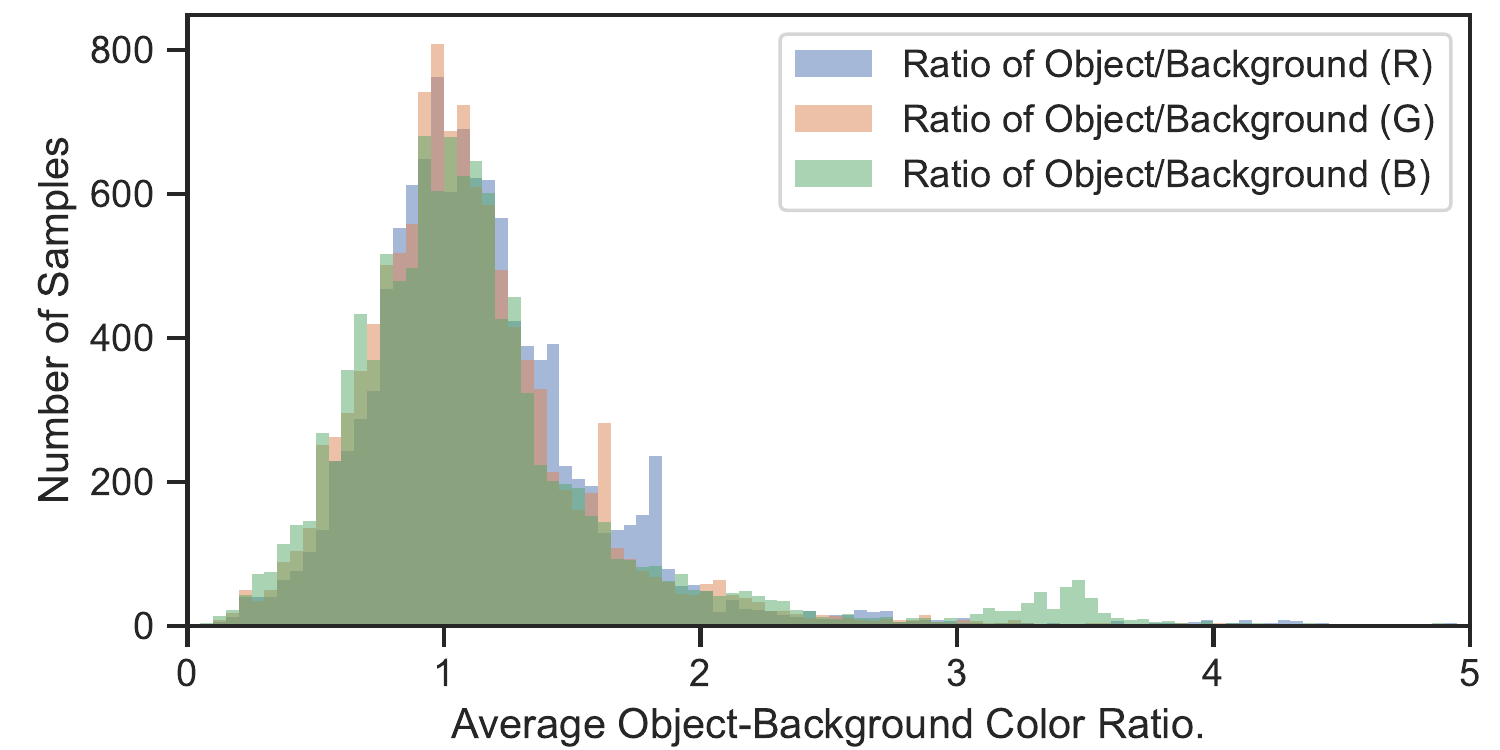}
      \caption{Average Color Ratio.}
      \label{fig:color_ratio}
    \end{subfigure}
    \par
    \begin{subfigure}{0.49\linewidth}
      \centering
      \includegraphics[width=\linewidth]{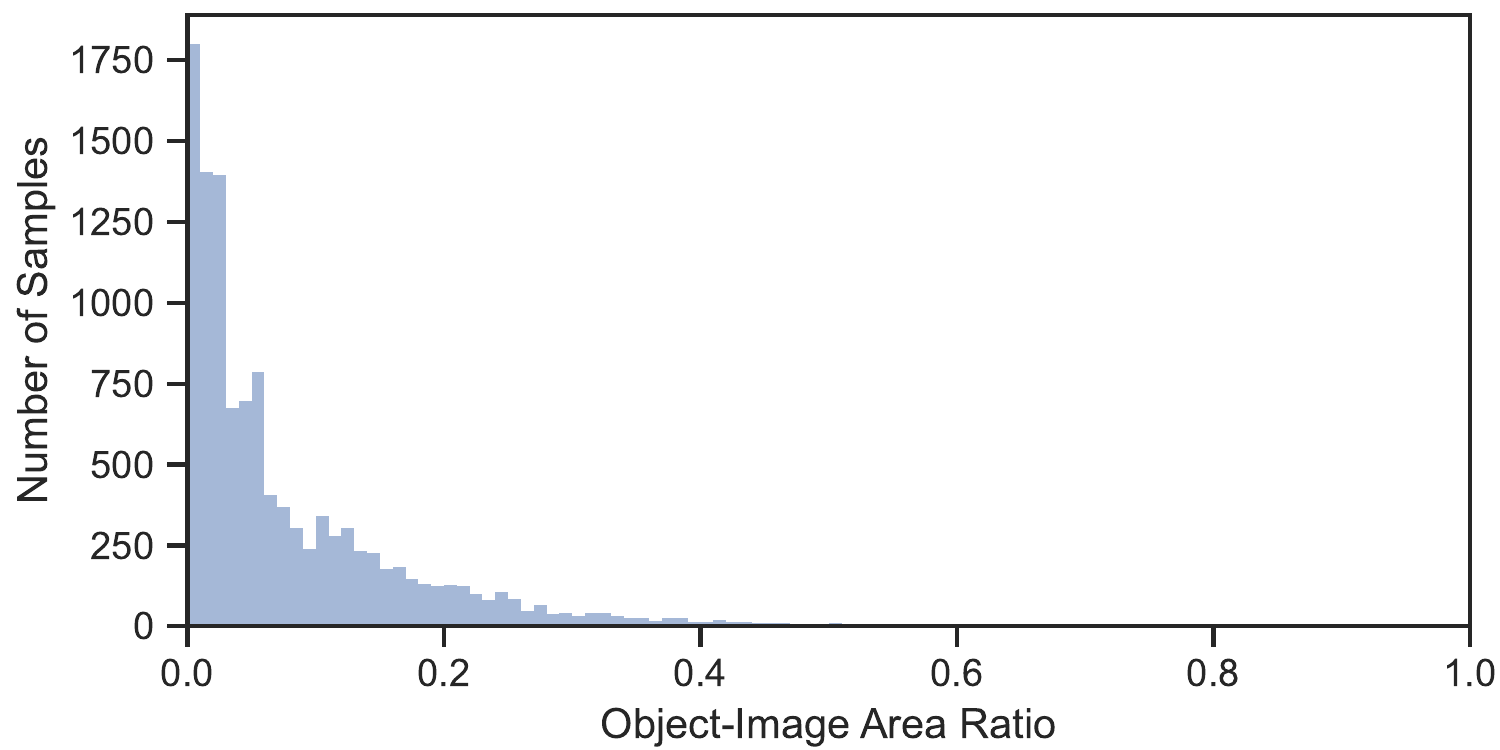}
      \caption{Object-Image Area Ratio.}
      \label{fig:object_area_ratio}
    \end{subfigure}
    \begin{subfigure}{0.49\linewidth}
      \centering
      \includegraphics[width=\linewidth]{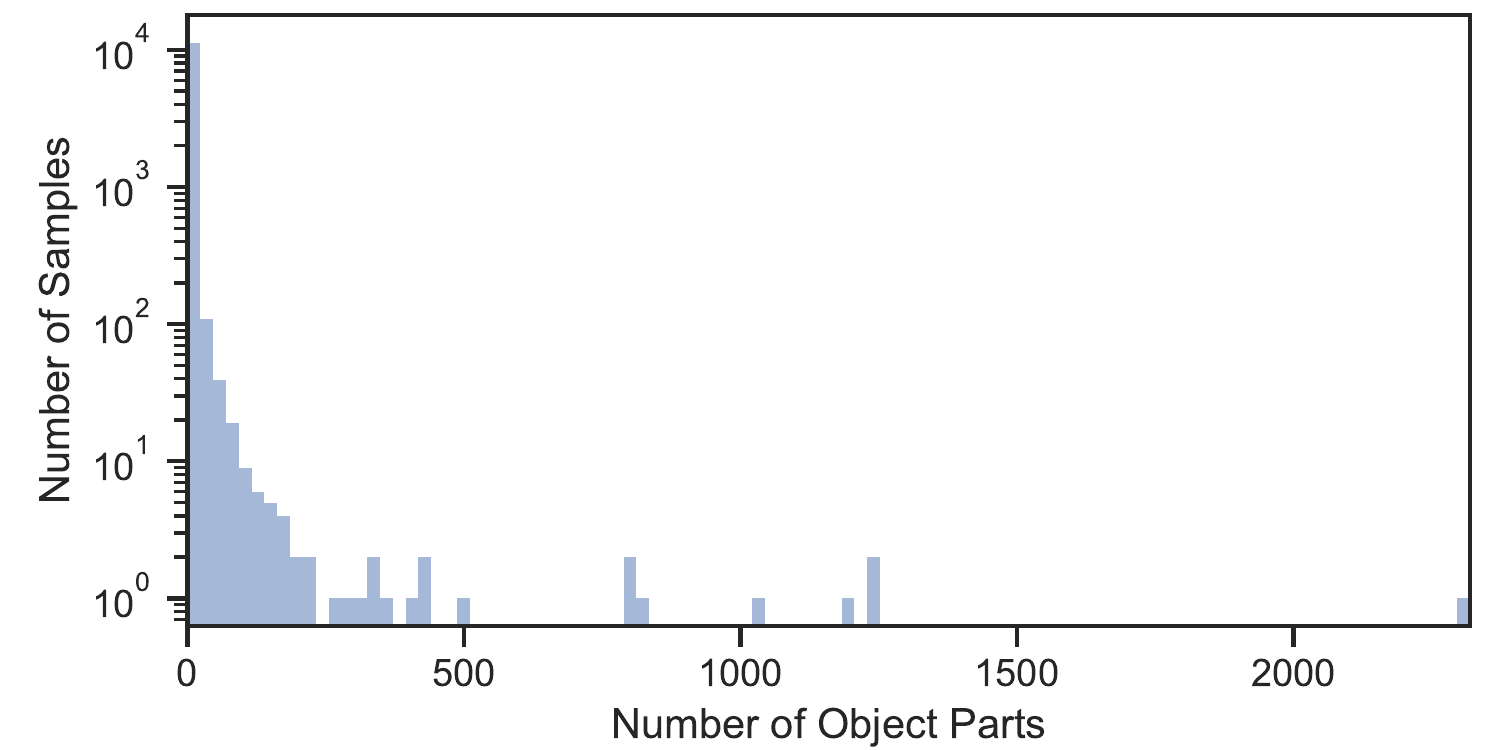}
      \caption{Number of Object Parts.}
      \label{fig:num_parts}
    \end{subfigure}
  \end{subfigure}
  \begin{subfigure}{0.34\linewidth}
    \centering
    \includegraphics[width=\linewidth]{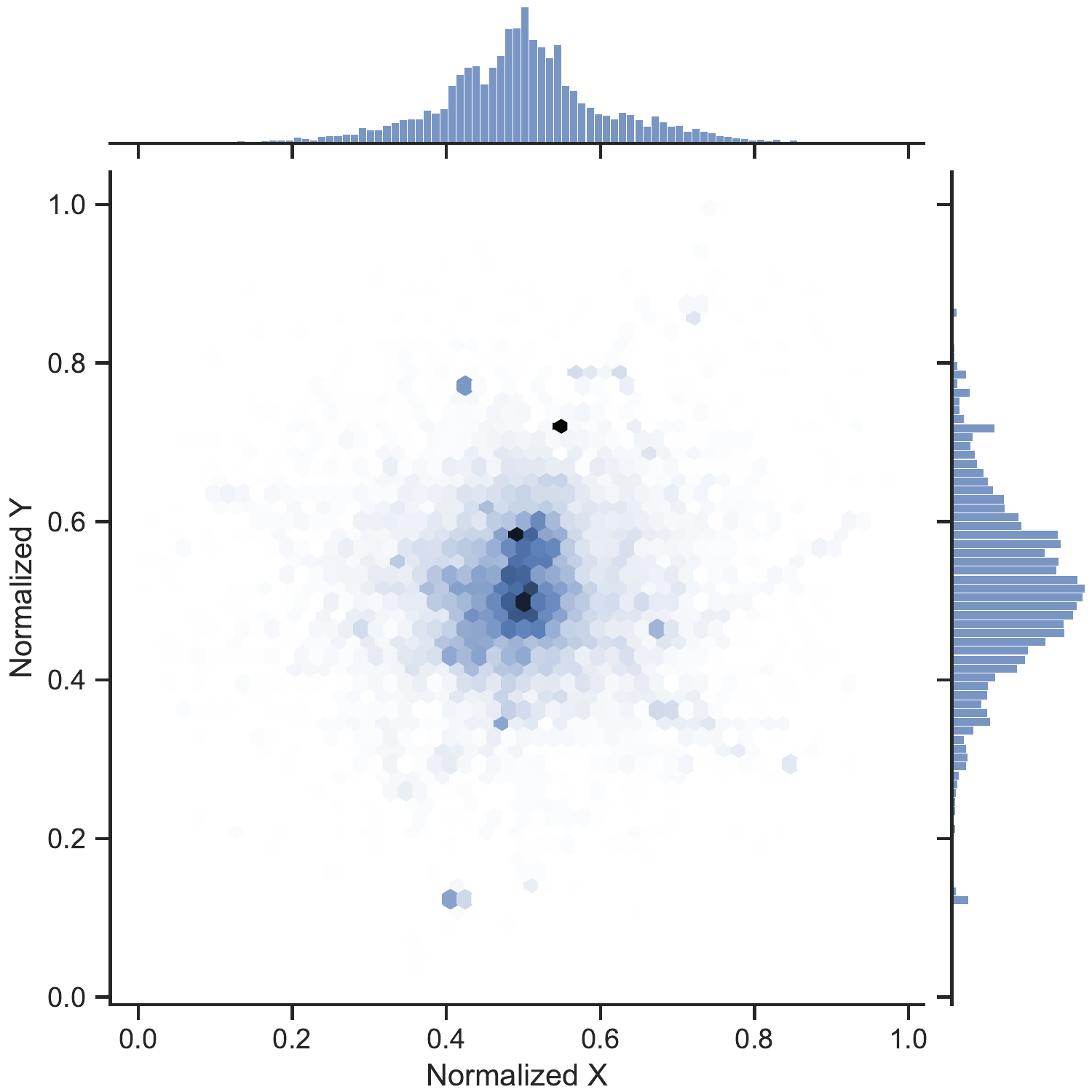}
    \caption{Normalized Centroid.}
    \label{fig:normalized_center}
  \end{subfigure}

  \caption{Attribute visualization of objects in our \mydataset dataset, including object concentration, object-background color ratio, object-image area ratio, number of object parts, and normalized centroid.
    Please refer to \cref{tab:dataattr} for details about attributes.}
  \label{fig:obj-attr}
  \vspace{-1em}
\end{figure}

\begin{wraptable}{r}{0.5\linewidth}
  \vspace{-3em}
  \caption{Attributes involved in the dataset analysis.}
  \resizebox{\linewidth}{!}{%
    \input{table/object_attribute.tex}
  }
  \label{tab:dataattr}
  \vspace{-2em}
\end{wraptable}
\parhead{Image Collection.}
Our data is collected from existing CSU datasets that have finely annotated segmentation maps.
Specifically, the \mydataset integrates 11,483 hand-selected images covering 75 object classes reconstructed from several public datasets~\cite{COD10K,CPD1K,PlantCamo,VCOD-CAD,VCOD-MoCA-Mask}.
The distribution of the number of samples in different classes is shown in~\cref{fig:num_classes}.
Meanwhile, we consider attributes of objects when selecting images, such as
object concentration,
average color ratio,
object-image area ratio,
number of object parts,
and normalized centroid.
\cref{tab:dataattr} gives their definitions.
And \cref{fig:obj-attr} visualizes the attribute distribution of the proposed dataset.
The camouflaged objects of interest usually have complex shape~\cref{fig:object_concertration},
high similarity to the background~\cref{fig:color_ratio}, and
small size~\cref{fig:object_area_ratio}.
And the image often contains multiple camouflaged objects or sub-regions with a central bias as shown in~\cref{fig:num_parts,fig:normalized_center}.

\parhead{Annotation.}
The original annotation cannot be used directly in the open-vocabulary setting, due to the following semantic ambiguities caused by different annotation standards, \ie,
1) Broad concepts, such as ``\texttt{fish}'' and ``\texttt{bug}'';
2) Vague definitions, such as ``\texttt{small fish}'' and ``\texttt{black cat}'';
3) Inconsistent granularity, such as the coexistence of ``\texttt{orchid mantis}'' and ``\texttt{mantis}'';
4) Non-entity concepts, such as ``\texttt{other}''.
These issues can lead to unreasonable and unreliable results for the open-vocabulary prediction as discussed in~\cite{OVSeg-OpenMetrics}.
To this end, we relabel the classes of all camouflaged objects and take the generality of the concept as the criterion for class definition, which also ensures lower semantic similarity.
The similarity analysis and rough hierarchy of the class set $\fontset{C}$ can be found in \cref{sec:dataset_details}.

\parhead{Data Division.}
To objectively evaluate the open-vocabulary segmentation algorithm on unseen classes, we assign as many classes as possible to the test set and control the sample ratio of the training set to the testing set to be 7:3.
Specifically, 14 classes in the dataset are taken as the training set and all the remaining 61 classes are used for testing.
Such a setting follows the existing practice in OVSIS where the number of seen classes (\eg, 171~\cite{COCO-Stuff}) is usually fewer than unseen classes (\eg, 847~\cite{ADE20K}) which is also closer to the real-world setting.
Such a setup can ensure the quantity of training samples while reinforcing the complexity of the test set.
Finally, the overall ratio of samples is 7713:3770.

\begin{figure}[t]
  \centering
  \includegraphics[width=\linewidth]{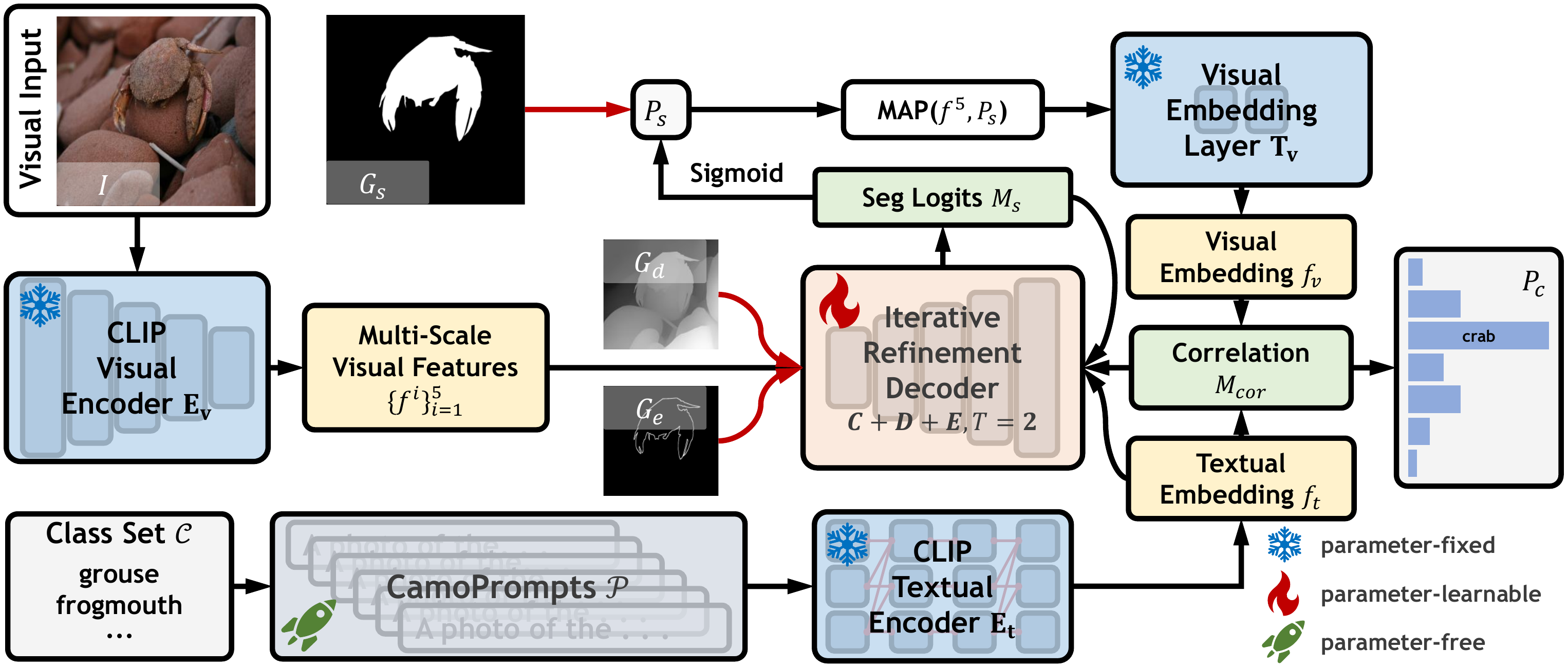}
  \caption{Overview of our single-stage open-vocabulary camouflaged object segmentation framework, \mymethod.
    It is based on the frozen CLIP model which includes feature encoder $\fontmod{E}_v$ and embedding layer $\fontmod{T}_v$ for visual appearance cues and textual encoder $\fontmod{E}_t$ for class semantic information.
    The well-constructed prompt template set \myprompt $\fontset{P}$ enables textual embedding $f_t$ to be more appropriate to \mytask.
    In the iterative refinement decoder (\cref{fig:isesg}), a more accurate class-related segmentation $P_{s}$, can be obtained with the assistance of the semantic guidance $\fontcomp{C}$ (\cref{fig:sga}) from the class semantic
    and the structure enhancement (\cref{fig:sea}) from the auxiliary depth and edge supervisions, \ie, $\fontcomp{D}$ and $\fontcomp{E}$.
    After being masked average pooled (MAP) by $P_{s}$, the high-level feature $f^5$ is used to assign a class $P_{c}$ to the object from the class set $\fontset{C}$.
    \textbf{Note that $\fontcomp{D}$ and $\fontcomp{E}$ are only used as auxiliary supervision during training and are not required during inference.}
  }
  \label{fig:net}
  \vspace{-1em}
\end{figure}

\section{Methodology}
\label{sec:methodology}

\begin{figure}[t]
  \centering
  \includegraphics[width=0.8\linewidth]{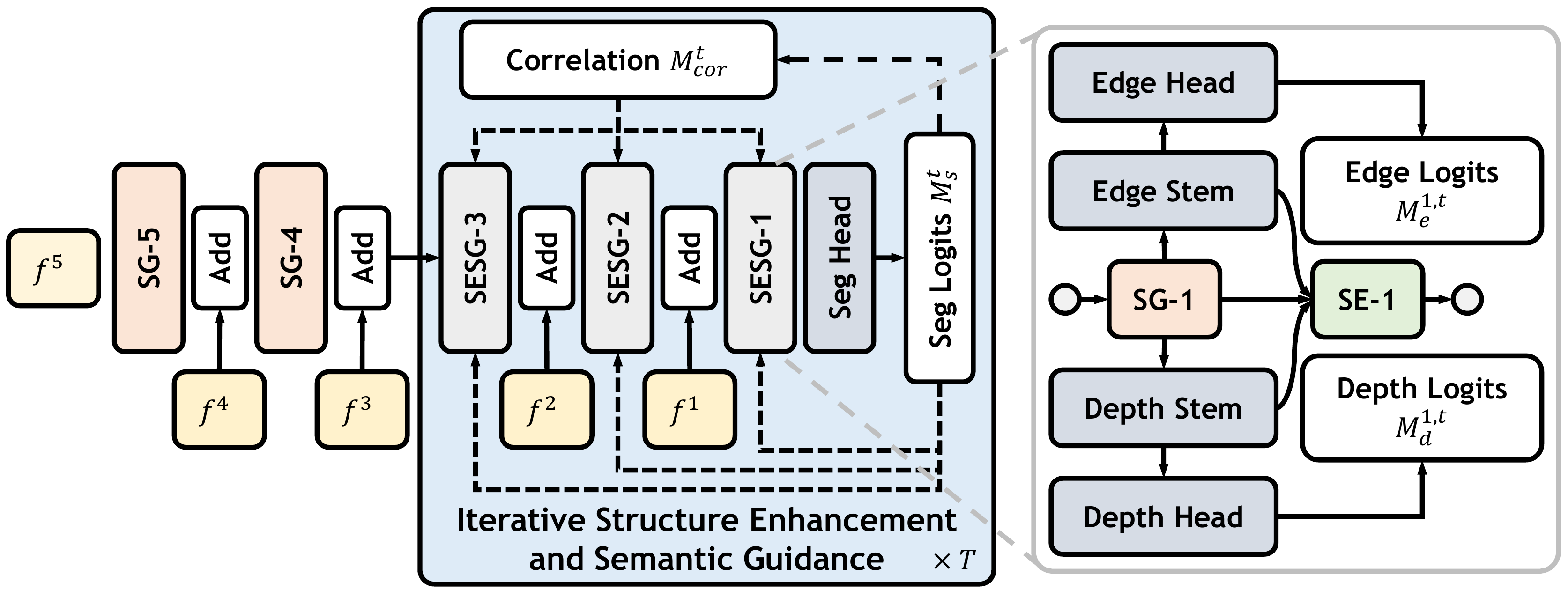}
  \caption{Proposed pipeline of the iterative refinement decoder with semantic guidance (SG) and structure enhancement (SE) components denoted as ``SG-$\star$'' and ``SE-$\star$''.
    $t$ and $T$ denote the current step and total number of iterations, respectively.
  }
  \label{fig:isesg}
  \vspace{-1em}
\end{figure}

In this section, we introduce a strong baseline \mymethod.
The overall framework is first described, followed by the details of the key components.

\parhead{Overall Architecture.}
We follow the common encoder-decoder paradigm and the pipeline is shown in~\cref{fig:net}.
Specifically, we first leverage the textual encoder $\mathbf{E_t}$ of the frozen CLIP to extract semantic embedding $f_t$ from the class label set $\fontset{C}$, and the visual encoder $\mathbf{E_v}$ to extract multi-scale image features $\{f^i\}^5_{i=1}$.
Both of them are fed into the decoder as the information bedrock of object segmentation, as shown in~\cref{fig:isesg}.
Structural cues such as depth and edge are also introduced to assist the iterative refinement process.
Finally, the class-related segmentation $P_s$ which is the segmentation logits $M_s$ after the \texttt{sigmoid} function processing, is used to remove the interference from the background in the high-level image feature and guide the generation of object-oriented visual representation $f_v$.
And the class label $P_{c}$ is determined by the similarity matching between $f_v$ and $f_t$.
More details about the proposed architecture are listed in \cref{sec:model_details}.

\begin{figure}[t]
  \centering
  \begin{subfigure}{0.49\linewidth}
    \centering
    \includegraphics[height=9em]{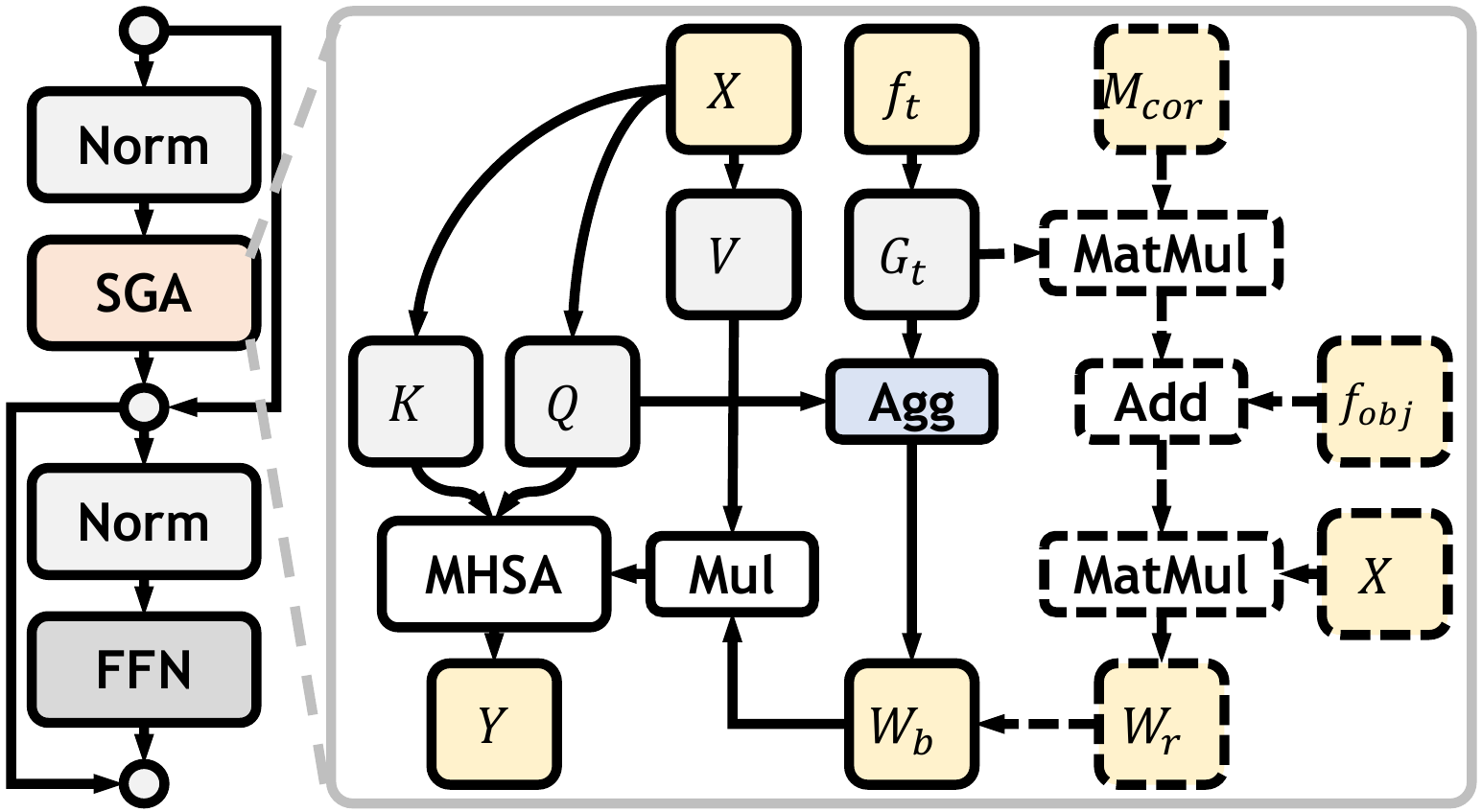}
    \caption{SG}
    \label{fig:sga}
  \end{subfigure}
  \hfill
  \begin{subfigure}{0.49\linewidth}
    \centering
    \includegraphics[height=9em]{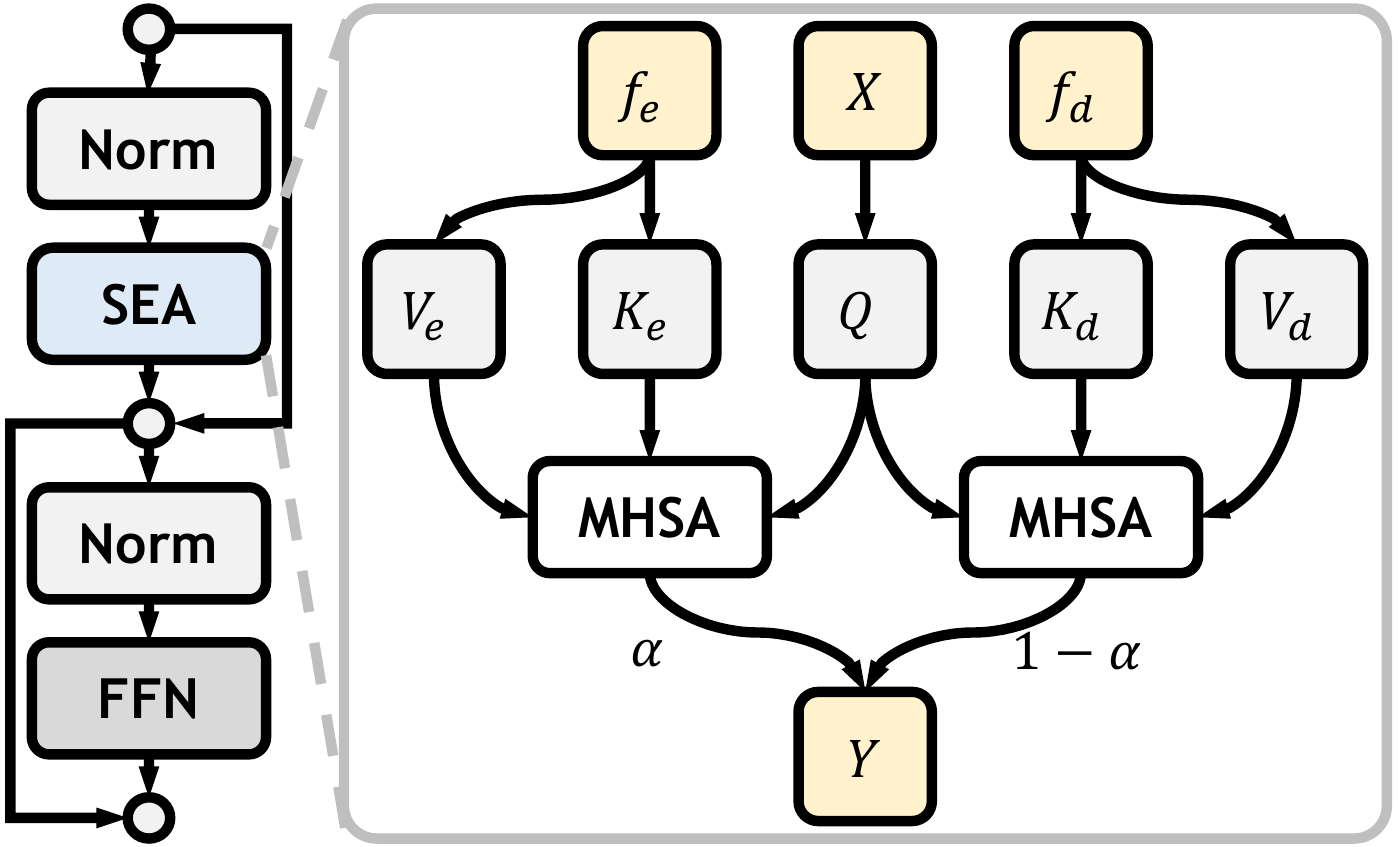}
    \caption{SE}
    \label{fig:sea}
  \end{subfigure}
  \caption{Proposed semantic guidance (SG) and structure enhancement (SE) components in which semantic guidance attention (SGA) and structure enhancement attention (SEA) are embedded. The dashed blocks represent the feedback paths that come into play during iteration.}
  \label{fig:attn}
  \vspace{-1em}
\end{figure}

\parhead{Semantic Guidance (SG).}
The class label definition usually is independent of the image scene.
It is very important to fully utilize the class prior for object recognition in complex scenes.
In the proposed decoder, normalized textual embedding is introduced into each stage to highlight semantically relevant cues.
As shown in~\cref{fig:sga}, we design a semantic guidance component SG to inject concept cues into the self-enhancement of the image feature.
Specifically, in our \textbf{semantic guidance attention (SGA)}, the normalized image feature $X$ is linearly mapped to $Q$, $K$, and $V$, while the textual embedding $f_t$ is transformed to the class guidance vector $G_t$.
The similarity between $Q$ and $G_t$ reflects the activation of different classes in spatial locations.
In \texttt{Agg} of~\cref{fig:sga}, the base weight $W_b$ for the spatial guidance is obtained after highlighting the most relevant class information by \texttt{softmax} operation.
And then $V$ is modulated and fed into \texttt{MHSA}, \ie, multi-head self-attention~\cite{Transformer}.

\parhead{Structure Enhancement (SE).}
Existing methods demonstrate that low-level structural information, such as the edge~\cite{COD-BGNet} and the depth~\cite{COD-Depth}, plays an important role in CSU, which is closely related to the mechanism of the human visual system.
So the SE component attached to the low-level SG is proposed to integrate the edge-aware and depth-aware cues and improve the structural details.
Specifically, the output of the SG is fed into two separate branches containing the convolutional stem and head for the edge and depth estimation as shown in~\cref{fig:isesg}.
The edge and depth logits maps, \ie, $M^{i}_e$ and $M^{i}_d$, from the head in the branch of the layer $i \in \{1, 2, 3\}$ are directly supervised.
And the outputs $f^i_e$ and $f^i_d$ of the stem are fed into the SE.
In the \parhead{structure enhancement attention (SEA)}, they independently update the normalized visual features $X$ using \texttt{MHSA}~\cite{Transformer}, and the corresponding outputs are combined with the learnable weight $\alpha$ as in~\cref{fig:sea}.

\parhead{Iterative Refinement.}
In the SG component, the aggregation process between image features and class semantics is not aligned and requires data-driven optimization.
Considering the aligned embedding space of the pre-trained CLIP, we introduce the correlation matrix $M_{cor}$ between the visual and textual embeddings into the SG as shown in~\cref{fig:sga}.
Meanwhile, due to the emphasis of the decoder output on the object region, the object-aware representation $f_{obj}$ is also inputted, which comes from the image features pooled by the coarse segmentation prediction in the last iteration.
By combining the two, we obtain task-oriented object cues, which is actually inspired by the top-down attention mechanism in the human cognitive system~\cite{BottomUpTopDownAttention,BottomUpTopDownAttentionVL}.
The spatial activation map $W_r$ of such object cues over image features is used to re-modulate $W_b$.
Besides, the SE in the iteration also helps the model to further optimize the texture details.
To benefit as much as possible from the assistance from structure enhancement while avoiding over-computation, we set the iteration entry to the third decoding layer as shown in~\cref{fig:isesg}.

\parhead{\myprompt.}
As mentioned in~\cite{CLIP}, prompt engineering and ensembling are important for the transfer performance of CLIP on downstream tasks, and the prompt template should be more relevant to the data type.
Because additional task-related cues are generally able to impose the necessary contextual constraint to the flexible CLIP.
Hence, instead of common practices~\cite{LearnablePrompt-CoOp,CLIP,OVOD-ViLD}, we design a simpler yet more effective template set \myprompt tailored for \mytask to decorate the class name, and average their textual embeddings as the final semantic embedding for each class.
Its full form is depicted in~\cref{tab:prompt}, while it also achieves better classification performance in comparison with other forms.

\parhead{Supervision.}
In each iteration, in addition to semantic object segmentation, we also need to perform depth estimation and edge estimation as auxiliary tasks.
For the segmentation prediction, we follow the commonly used weighted segmentation loss function $l^{t}_{s}=l_{s}(P^t_{s},G_{s})$~\cite{COD-CamoFormer,COD-BGNet}.
For the edge estimation, considering the imbalance problem of positive and negative samples, we introduce the dice loss function as $l^{i,t}_{e}=l_{e}(P^{i,t}_{e},G_{e})$.
The summation of L1 and SSIM losses, \ie, $l^{i,t}_{d}=l_{d}(P^{i,t}_{d},G_{d})$, is used for the depth estimation.
The total loss $L$ of our method can be formulated as follows:
$L = \sum^{T}_{t=1} \left [ l^{t}_{s} + \sum^{3}_{i=1} \left ( l^{i,t}_{e} + l^{i,t}_{d} \right ) \right ]$,
where $t$ and $i$ are used to index iterations and layers, respectively.
And the total number $T$ of iterations is set to 2 as mentioned in~\cref{sec:ablation}.

\begin{figure}[t]
  \centering
  \includegraphics[width=\linewidth]{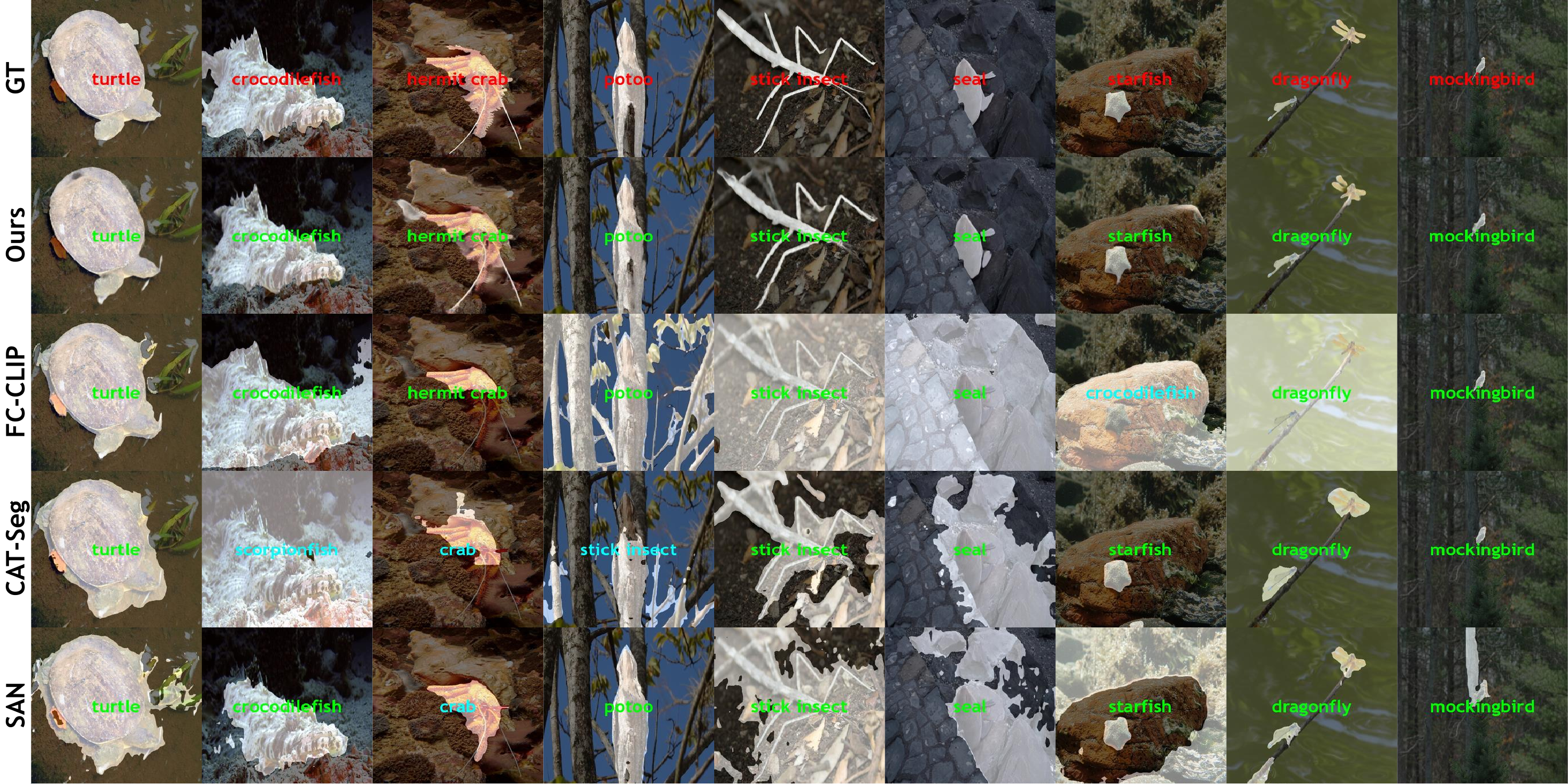}
  \caption{Visual results on \mydataset.
    Existing methods are either disrupted by chaotic backgrounds, imperceptible appearances, blurry details, or severe occlusion, while our algorithm can effectively capture and remain well-exposed object details.
    And three different colors are used to represent \textcolor{red}{human annotations}, \textcolor{truecls}{correct predictions}, and \textcolor{falsecls}{incorrect predictions}.
  }
  \label{fig:viscmp}
  \vspace{-1em}
\end{figure}

\section{Experiments}
\label{sec:experiments}

\subsection{Implementation Details}

\parhead{Dataset Settings.}
As mentioned in~\cref{sec:dataset}, the proposed dataset is divided according to classes into two disjoint subsets, \ie, $\fontset{C}_{seen}$ and $\fontset{C}_{unseen}$.
The former contains 14 classes for training and the latter contains the remaining 61 classes for testing.
The proposed dataset itself is provided with only images and masks.
We use the typical monocular depth estimation methods~\cite{DepthEstimation-DPT,DepthEstimation-MaskDepth} to obtain the depth map $G_{d}$ for training, while the edge map $G_{e}$ is generated by dilating and eroding operations.
\textit{We introduce depth and edge maps only in the training phase and they are not required during inference, which also avoids information leakage during testing.}
So, our depth and edge maps are labor-free and training-only, and directly generated by existing algorithms.
Such a design can help our model get rid of the dependence on them in real-world applications.
Our structure enhancement based on depth and edge information belongs to a deeper exploration of this task.
These generated depth maps will be made publicly available with the dataset.

\parhead{Training Details.}
Following previous settings in~\cite{OVSeg-FCCLIP,OVSeg-SimSeg,OVSeg-SAN}, the pre-trained CLIP is frozen during training, and the remaining parameters are learnable and are randomly initialized.
The AdamW~\cite{AdamW-SGDW} optimizer with the learning rate of 3e-6, weight decay of 5e-4, batch size of 4, and training epoch of 30 is used to optimize model parameters.
Some basic data augmentations including random flipping, rotating, and color jittering, are introduced to preprocess training data.
The input data and the output logits are bilinearly interpolated to $384 \times 384$ during training and inference.

\parhead{Evaluation Protocol.}
To reasonably evaluate the performance of \mytask, we modify the metrics in the original CSU task~\cite{SINetV2,VCOD-MoCA-Mask} to
cS$_m$, cF$^{\omega}_{\beta}$, cMAE, cF$_{\beta}$, cE$_{m}$, and cIoU
which follows the common settings in the OVSIS field~\cite{OVSeg-SimSeg,OVSeg-OVSeg,OVSeg-SAN,OVSeg-CATSeg,OVSeg-FCCLIP,Survey-OpenVocabulary} and takes into account both classification and segmentation.

\subsection{System-level Comparison}

To show the complexity of the proposed \mytask task and also to verify the effectiveness of the proposed method, we compare \mymethod with several recent state-of-the-art methods in OVSIS, including~\cite{OVSeg-SimSeg,OVSeg-OVSeg,OVSeg-ODISE,OVSeg-SAN,OVSeg-CATSeg,OVSeg-FCCLIP}.
Since this is a new task, existing methods need to be re-evaluated to understand their generalization ability.
Based on the public code and weights trained on COCO-Stuff~\cite{COCO-Stuff} provided by the authors, we show the performance of these methods under three different testing schemes, including
S.I) testing directly with their trained weights;
S.II) further fine-tuning based on trained weights before testing;
and S.III) testing after re-training directly on our training set.

\parhead{Quantitative Evaluation.}
For the sake of fairness in comparisons, we report the performance of the ``Large'' versions for these methods, except for SimSeg~\cite{OVSeg-SimSeg} where the authors only provide the ``Base'' version.
All results are summarized in~\cref{tab:cmpsota} and our approach consistently outperforms these competitors.
It is worth noting that existing methods perform better at S.I.
This may be attributed to the training process on the larger-scale COCO-Stuff dataset, which provides a more general understanding of the concepts.
However, direct fine-tuning (\ie, S.II) may destroy this knowledge and even cause oblivion to some extent, resulting in performance degradation.
If we follow S.III to re-train them, the relatively small-scale training data may not be enough to train these models with more complex structures.
At the same time, the existing methods also lack targeted optimization for the \mytask task.
These problems can lead to further deterioration of performance.
However, our approach tailored for \mytask achieves leading performance by the iterative refinement strategy of multi-source information, which comprehensively considers different characteristics of the task.

\parhead{Qualitative Evaluation.}
We also visualize the results of some recent methods on a variety of data in~\cref{fig:viscmp}.
It can be seen that the proposed method shows better performance and adaptability to diverse objects, including
large objects (Col. 1-2),
middle objects (Col. 3-5),
small objects (Col. 6-9),
multiple objects (Col. 8),
complex shapes (Col. 3-5),
blurred edges (Col. 1-5),
severe occlusion (Col. 6),
and background interference (Col. 2-6).

\begin{table}[t]
  \centering
  \caption{Comparison with recent state-of-the-art CLIP-based open-vocabulary semantic image segmentation methods with different training settings on the proposed \mydataset dataset.
    The best three results are highlighted in {\color{reda} \textbf{red}}, {\color{mygreen} \textbf{green}} and {\color{myblue} \textbf{blue}}.
  }
  \resizebox{\linewidth}{!}{%
    \input{table/cmpsota_simple.tex}
  }
  \label{tab:cmpsota}
  \vspace{-1em}
\end{table}

\begin{table}[t]
  \centering
  \caption{Ablation comparison of proposed components.
    $\Delta$ represents the \textit{average relative gain} in performance of the corresponding model over the baseline for \mytask.
    $\fontset{P}$: \myprompt.
    $\fontcomp{C}$: Semantic guidance.
    $\fontcomp{D}$: Depth estimation auxiliary task.
    $\fontcomp{E}$: Edge estimation auxiliary task.
    $T$: Number of iterations which is set to 2 by default due to the best performance.
    ``$\lim\sup_{P_{s} \rightarrow G_{s}}$Perf.'': The ideal performance for our framework.
  }
  \resizebox{0.8\linewidth}{!}{%
    \input{table/ablation.tex}
  }
  \label{tab:ablation}
\end{table}

\subsection{Analysis and Ablation Study}
\label{sec:ablation}

\parhead{Better Baseline Model.}
Interestingly, our baseline model in~\cref{tab:ablation} outperforms existing methods in~\cref{tab:cmpsota}.
We attribute this to the fact that the overly complex feature pipeline of the conventional OVSIS architecture is not well adapted to this task.
And our framework, from hyperparameter setting to structural evolution, is explored for the proposed task and data.
Thus, we actually provide a more effective and robust baseline setup for \mytask.

\parhead{Importance of Modules.}
To specifically analyze the effect of different components, we evaluate their performance in~\cref{tab:ablation}.
As can be seen, the proposed modules all show positive gains.
Both the semantic guidance in the class-aware decoding and the structure enhancement from auxiliary tasks of depth and edge estimation consistently boost the final performance.
The ablation comparison also demonstrates that explicit guidance of the spatial and contour information is important for the detection of camouflaged objects and their positive gains are subject to the effective fusion design.
We replace the proposed structure enhancement (SE) component with the vanilla addition fusion with a similar number of parameters, which results in detrimental effects.
Besides, in~\cref{tab:ablation}, we also give the ideal results of our CLIP-driven framework where $G_s$ is treated as $P_s$.
Although our method exhibits leading performance compared to existing methods as shown in~\cref{tab:cmpsota}, there is still a long way to go to solve this problem.
At the same time, \textit{the current ideal performance is still far from the limit, suggesting that future breakthroughs in this field may require more powerful paradigms} as discussed in \cref{sec:limitation}.

\parhead{Importance of Iterative Refinement.}
The top-down iterative refinement strategy significantly improves the \mytask performance as shown in~\cref{tab:ablation}.
When the number $T$ of iterations is 2, our algorithm obtains the best \mytask performance.
As $T$ increases, there is no further improvement in performance, so it is set to 2 by default.
In addition, the object-aware representation $f_{obj}$ from the output space plays an important role,
and introducing the correlation guidance $M_{cor}$ also has positive gains.

\begin{wrapfigure}{r}{0.50\linewidth}
  \vspace{0em}
  \centering
  \includegraphics[width=\linewidth]{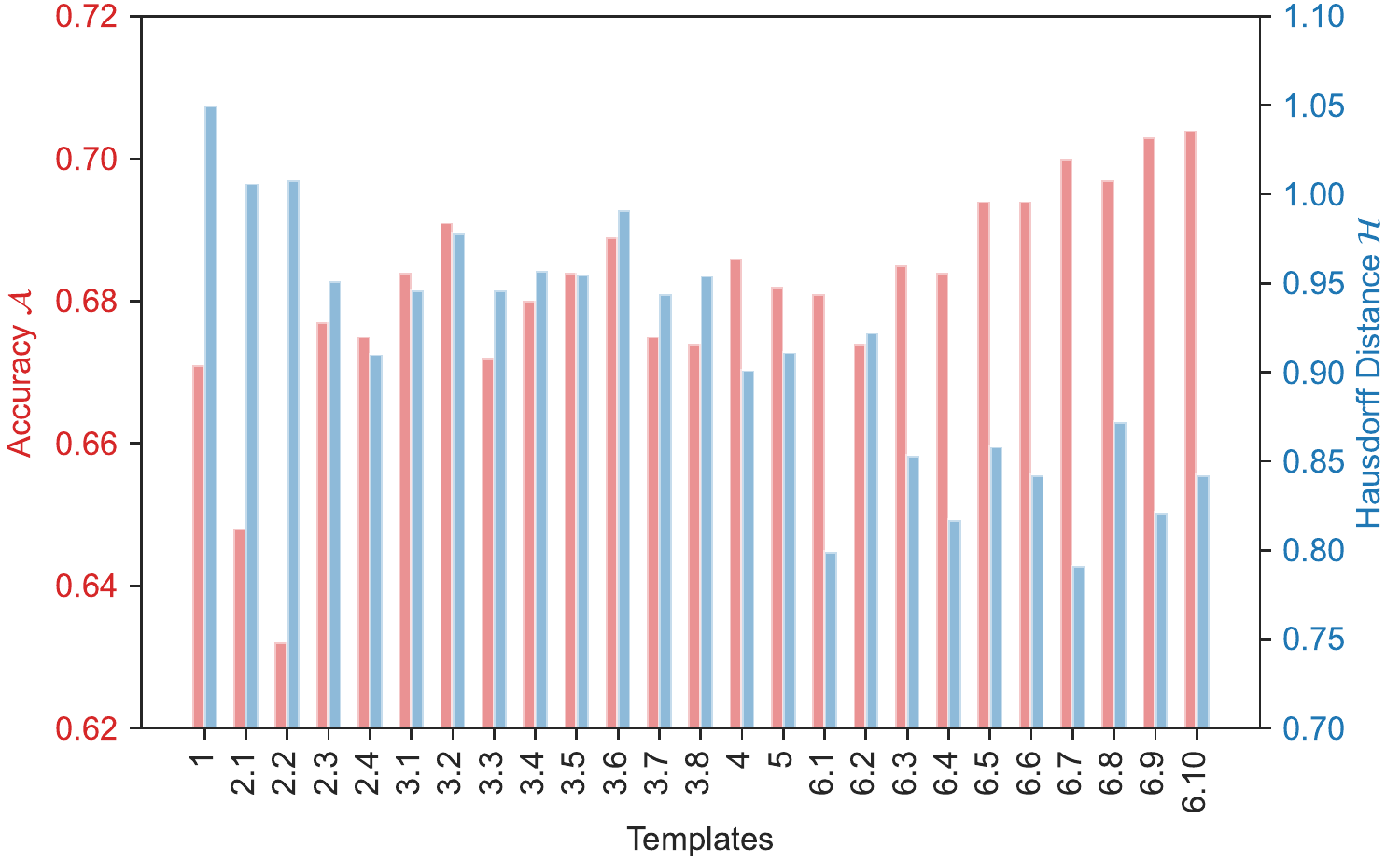}
  \caption{Illustration of classification accuracy $\mathcal{A}$ and Hausdorff distance $\mathcal{H}$ between the textual embeddings of training and testing class splits by using the plain CLIP on the \mydataset dataset.
    The approximate negative correlation between them may reveal that the templates that bring training and testing class embeddings closer usually also improve classification accuracy.
    \cref{tab:prompt} shows the details of these templates.
  }
  \label{fig:prompt}
  \vspace{-2em}
\end{wrapfigure}
\parhead{Importance of \myprompt.}
Our prompt template set \myprompt, which takes task attributes into account, shows better performance in~\cref{tab:prompt}.
To further understand the influence of different templates on the semantic embedding, we calculate the Hausdorff distance between training and testing class labels in the embedding space, as shown in~\cref{fig:prompt}.
The figure presents an interesting phenomenon that those templates with better classification performance tend to reduce the distance, which inspires further explorations for more effective prompt engineering.

\begin{figure}[t]
  \captionof{table}{Classification accuracy $\mathcal{A}$ of the plain CLIP using different prompt templates on \mydataset, which is based on the masked average pooling with the ground truth mask.}
  \begin{minipage}{0.49\linewidth}
    \centering
    \resizebox{\linewidth}{!}{%
      \input{table/prompt_wmask1.tex}
    }
  \end{minipage}
  \hfill
  \begin{minipage}{0.49\linewidth}
    \centering
    \resizebox{\linewidth}{!}{%
      \input{table/prompt_wmask2.tex}
    }
  \end{minipage}
  \label{tab:prompt}
  \vspace{-1em}
\end{figure}

\begin{figure}[t]
  \begin{minipage}{0.61\linewidth}
    \centering
    \captionof{table}{Efficiency comparison with other methods.
      ``Trainable Param.'' and ``Total Param.'' stand for the number of trainable and total parameters.
      In OVSeg~\cite{OVSeg-OVSeg}, the CLIP~\cite{CLIP} model is fine-tuned by the authors, resulting in more trainable parameters.}
    \resizebox{\linewidth}{!}{%
      \input{table/efficiency.tex}
    }
    \label{tab:efficiency}
  \end{minipage}
  \hfill
  \begin{minipage}{0.38\linewidth}
    \vspace{1em}
    \centering
    \includegraphics[width=0.9\linewidth]{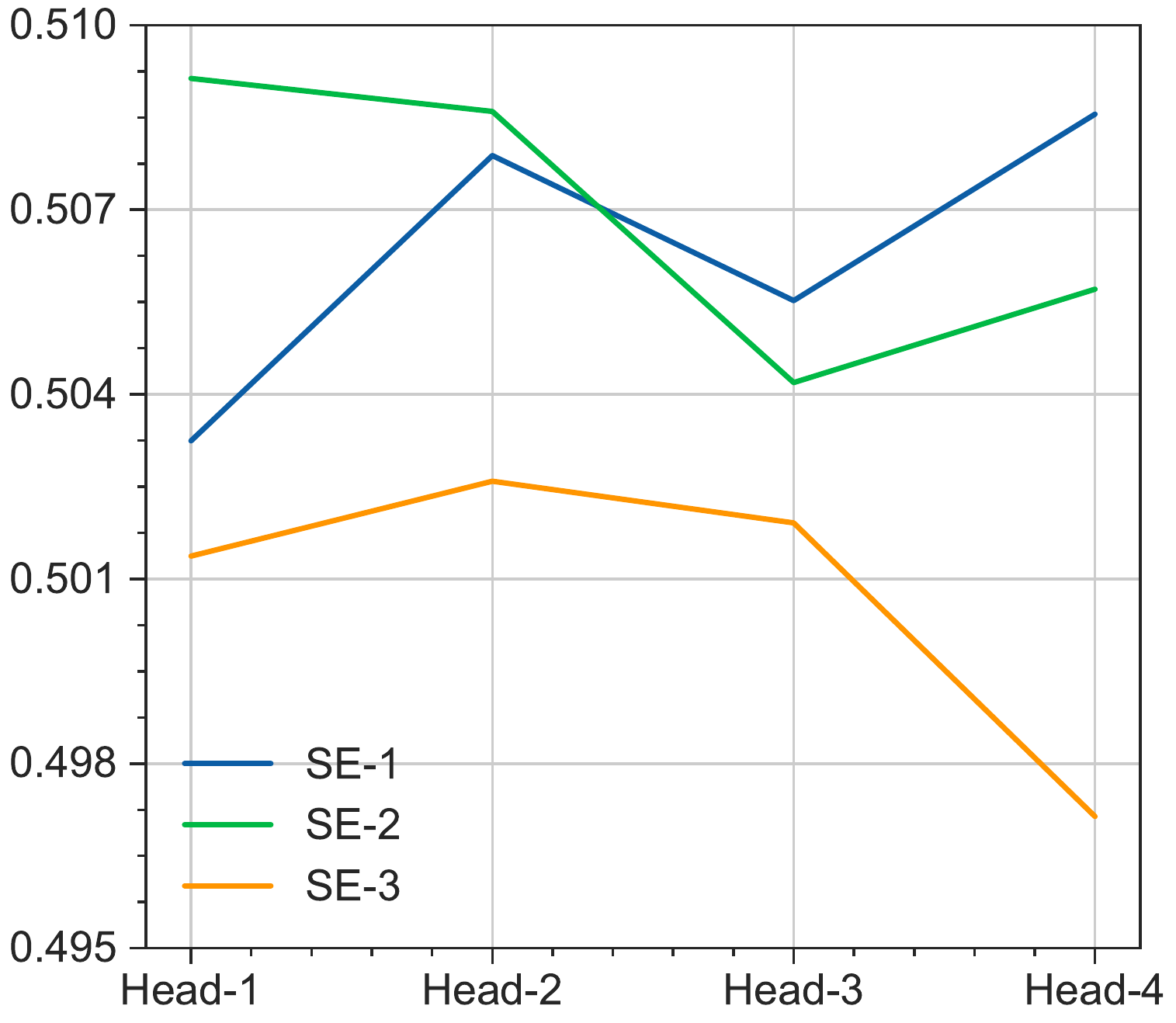}
    \captionof{figure}{$\alpha$ corresponding to different heads in~\cref{fig:sea} from different decoding layers.}
    \label{fig:alpha}
  \end{minipage}
\end{figure}

\parhead{Importance Between Edge and Depth.}
In the proposed SE as shown in~\cref{fig:sea}, the interaction components for the edge and depth are combined by a coefficient vector $\alpha$, which can also reflect the relative importance of the two kinds of information.
In~\cref{fig:alpha}, we plot  $\alpha$ from different decoding layers.
It can be seen that the values are usually greater than 0.5, which indicates a preference for the edge information flow.

\parhead{Model Complexity.}
We compare the proposed method with the CLIP-based competitors~\cite{OVSeg-SimSeg,OVSeg-OVSeg,OVSeg-ODISE,OVSeg-SAN,OVSeg-CATSeg,OVSeg-FCCLIP} in terms of the number of trainable and total parameters, and FLOPs.
To be fair, we test all methods following the setting of their original inference settings.
As can be seen from~\cref{tab:efficiency}, our method has fewer trainable parameters ($< 2\%$) and less computational complexity ($0.2T$), which is superior to these competitors.

\section{Conclusion}
\label{sec:conclusion}

In this work, we propose a new challenging task, \mytask, to explore open-vocabulary semantic image segmentation (OVSIS) for the camouflaged objects in more complex natural scenes, and carefully collect and construct a large-scale data benchmark \mydataset.
Meanwhile, by considering the characteristics of the task and data, we propose a strong single-stage baseline \mymethod with the advanced pre-trained vision-language model.
Specifically, the well-designed prompt templates are introduced to reinforce the task-relevant semantic context.
We introduce additional multi-source information including class semantic cues, depth spatial structure, object edge details, and top-down iterative guidance from the output space.
With the help of these components, \mymethod can perceive and segment camouflaged objects in complex environments.
Extensive experiments demonstrate the effectiveness of the proposed method and its superior performance compared with the existing state-of-the-art OVSIS algorithms on \mydataset.

{\footnotesize
\parhead{Acknowledgements.}
We sincerely thank these authors of the public datasets~\cite{COD10K,CPD1K,PlantCamo,VCOD-CAD,VCOD-MoCA-Mask} used in this work:
Deng-Ping Fan, Ge-Peng Ji, Guolei Sun, Ming-Ming Cheng, Jianbing Shen, and Ling Shao for COD10K~\cite{COD10K},
Yunfei Zheng, Xiongwei Zhang, Feng Wang, Tieyong Cao, Meng Sun, and Xiaobing Wang for CPD1K~\cite{CPD1K},
Jinyu Yang for PlantCamo~\cite{PlantCamo},
Pia Bideau and Erik Learned-Miller for CAD~\cite{VCOD-CAD},
and Xuelian Cheng, Huan Xiong, Deng-Ping Fan, Yiran Zhong, Mehrtash Harandi, Tom Drummond, and Zongyuan Ge for MoCA-Mask~\cite{VCOD-MoCA-Mask}.
It is their efforts that underpin this new exploration of camouflaged object segmentation.}

\appendix

\section{Model Details}
\label{sec:model_details}

\subsection{Details of $\fontmod{E}_v$ and $\fontmod{T}_v$}
In the proposed model, the visual encoder $\fontmod{E}_v$ and embedding layer $\fontmod{T}_v$ together are used to extract the high-level embedding corresponding to the object of interest in the input image.
The two independent sub-networks are split from the visual network of CLIP~\cite{CLIP,OpenCLIP}.
$\fontmod{E}_v$ contains all the feature encoding layers for extracting the multi-scale image features.
And $\fontmod{T}_v$ corresponds to the final high-dimensional projection layer, which is used to convert the high-level image feature $f^5$ into the visual embedding vector $f_v$.

\subsection{Multi-scale Image Features $\{f^i\}^5_{i=1}$}
The multi-scale image features $\{f^i\}^5_{i=2}$ is from the four stages of ConvNeXt~\cite{ConvNeXt} with different output resolutions, respectively.
While the feature $f^1$ is obtained by up-sampling the feature $f^2$ $2\times$ by bilinear interpolation.

\subsection{Class Embedding}
In the proposed algorithm, for each image input, the textual encoder needs to extract text embedding for all class texts.
However, due to the nature of the algorithm design, these text embeddings are shared in each iteration, so they can be pre-computed in the inference to save inference cost.

\subsection{Details of Classification}
During the inference, the class prediction $P_{c}$ is generated from the pair-wise correlation matrix $M_{cor}$ after the softmax operation, where $M_{cor}$ is from the multiplication between the normalized visual embedding and textual embedding, \ie $f_v$ and $f_t$.
Besides, in our experiments, introducing classification supervision on top of the existing form interferes with the training process of the model, which brings about significant performance degradation, with relative reductions of 8.6\% and 15.7\% for cS$_m$ and cF$^{\omega}_{\beta}$, respectively.
Therefore, we do not consider classification loss during training, and $M_{cor}$ is only used to construct the top-down iterative guidance.

\section{Dataset Details}
\label{sec:dataset_details}

\begin{figure}[h]
  \centering
  \begin{subfigure}{0.54\linewidth}
    \centering
    \includegraphics[width=\linewidth]{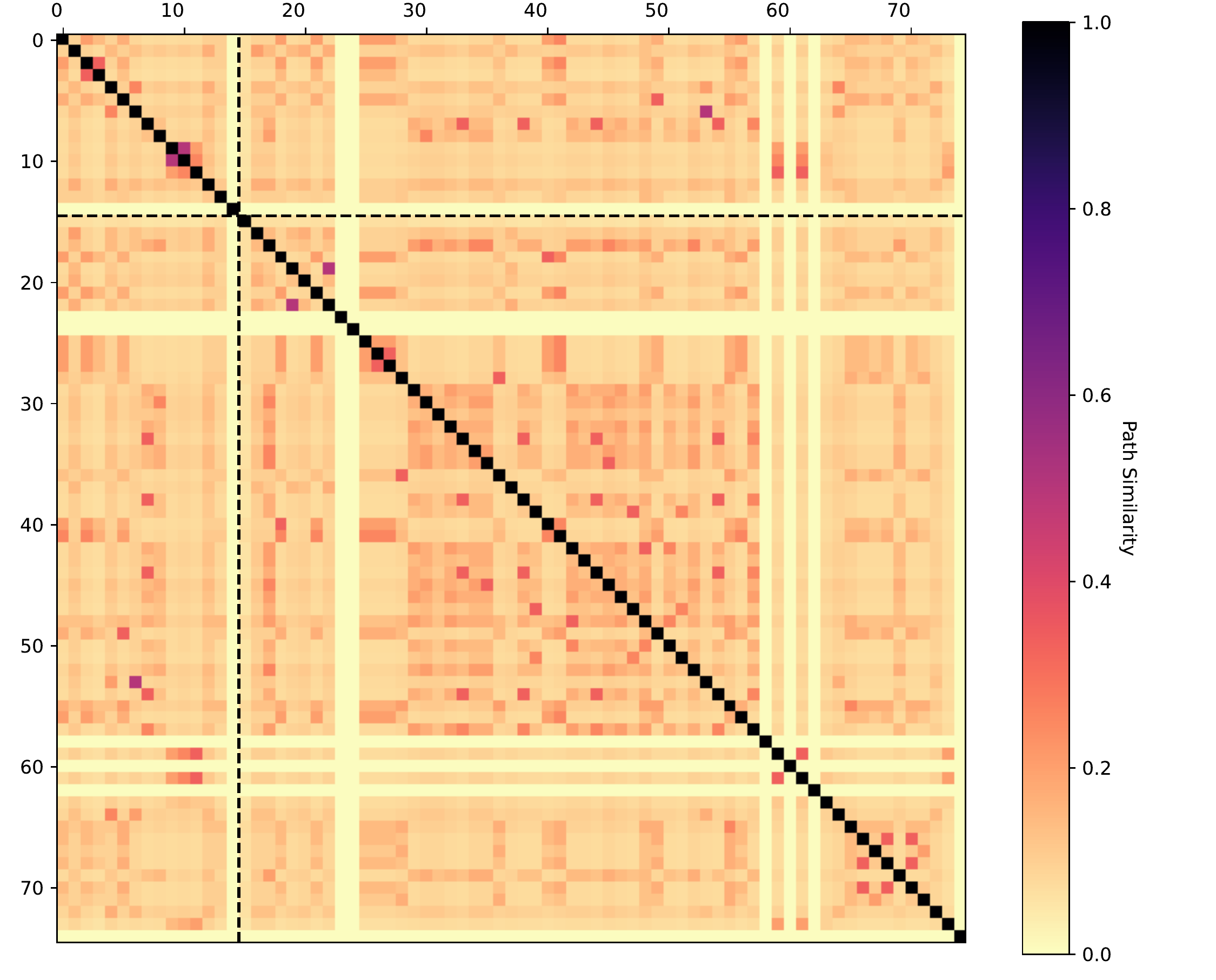}
    \caption{Semantic similarity score map.}
    \label{fig:simmap}
  \end{subfigure}
  \begin{subfigure}{0.43\linewidth}
    \centering
    \includegraphics[width=\linewidth]{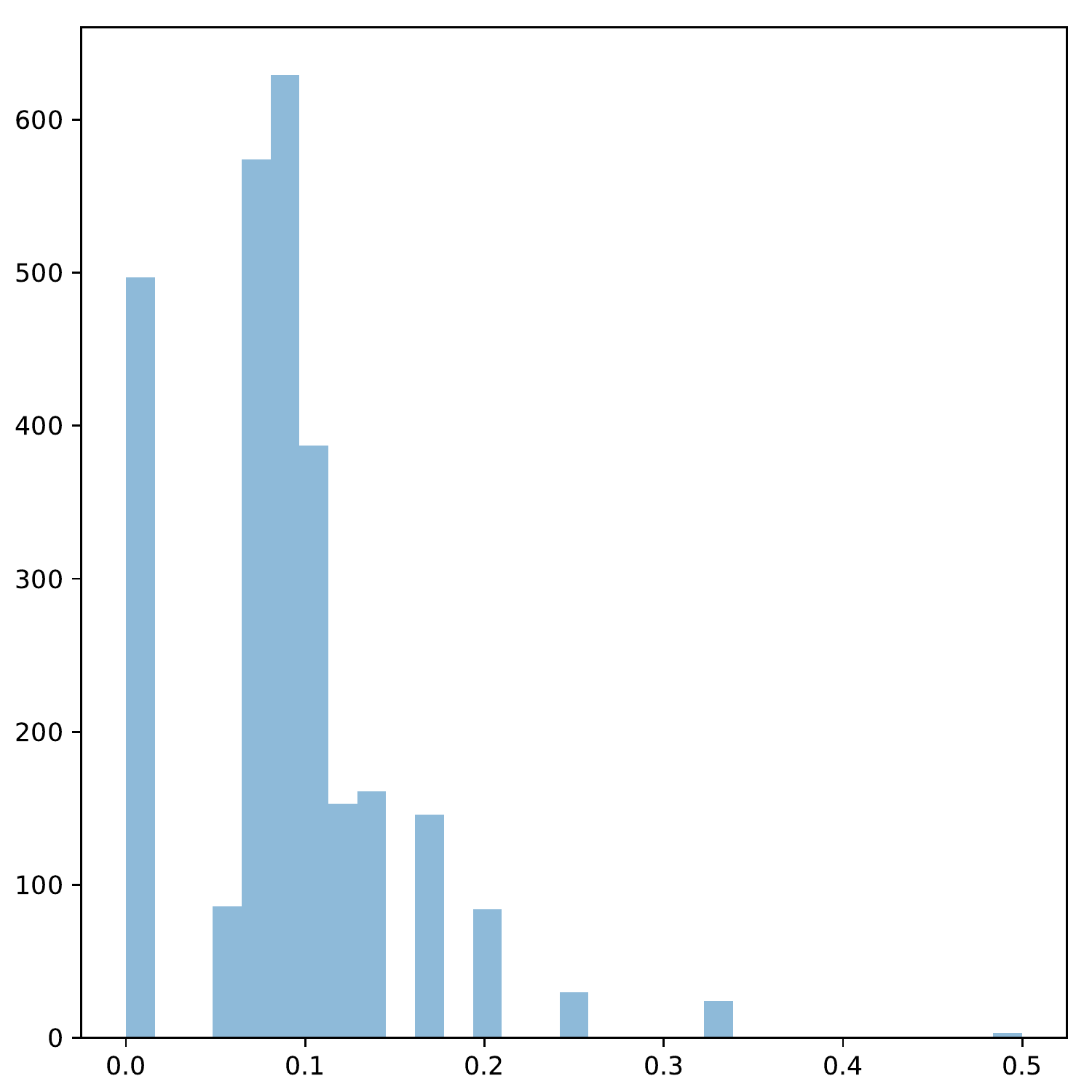}
    \caption{Semantic similarity score histogram.}
    \label{fig:simhist}
  \end{subfigure}

  \caption{Class semantic similarity of \mydataset based on the Open English WordNet~\cite{OpenEnglishWordNet}.
    The classes belonging to the training and testing sets are separated here using black dashed lines in (a).
    Note that only the similarity between two different classes is considered in (b).
  }
  \label{fig:similarity}
\end{figure}

\subsection{Class Semantic Similarity}
To analyze the semantic similarity between the relabelled classes, we compute the path similarity between the classes based on the Open English WordNet~\cite{OpenEnglishWordNet} as shown in~\cref{fig:similarity}.
Specifically, when $p$ is the length of the shortest path between two classes, the path similarity score $s$ is:
\begin{equation}
  s = \frac{1}{p+1}
\end{equation}
The score $s$ ranges between 0.0 and 1.0, where the higher the score is, the more similar the two classes are.
The score $s$ is 1.0 when a class is compared to itself, and 0.0 when there is no path between the two classes (i.e., the path distance is infinite).
As shown in~\cref{fig:simhist}, the semantic similarity between classes in our class set $\fontset{C}$ is very low.
Most of them lie around 0.1, while the maximum is only 0.5.
Such low similarity can better alleviate the complexity due to class semantic similarity during open vocabulary evaluation.

\subsection{Class Hierarchy Relationships}
\begin{figure}[h]
  \centering
  \includegraphics[width=0.4\linewidth]{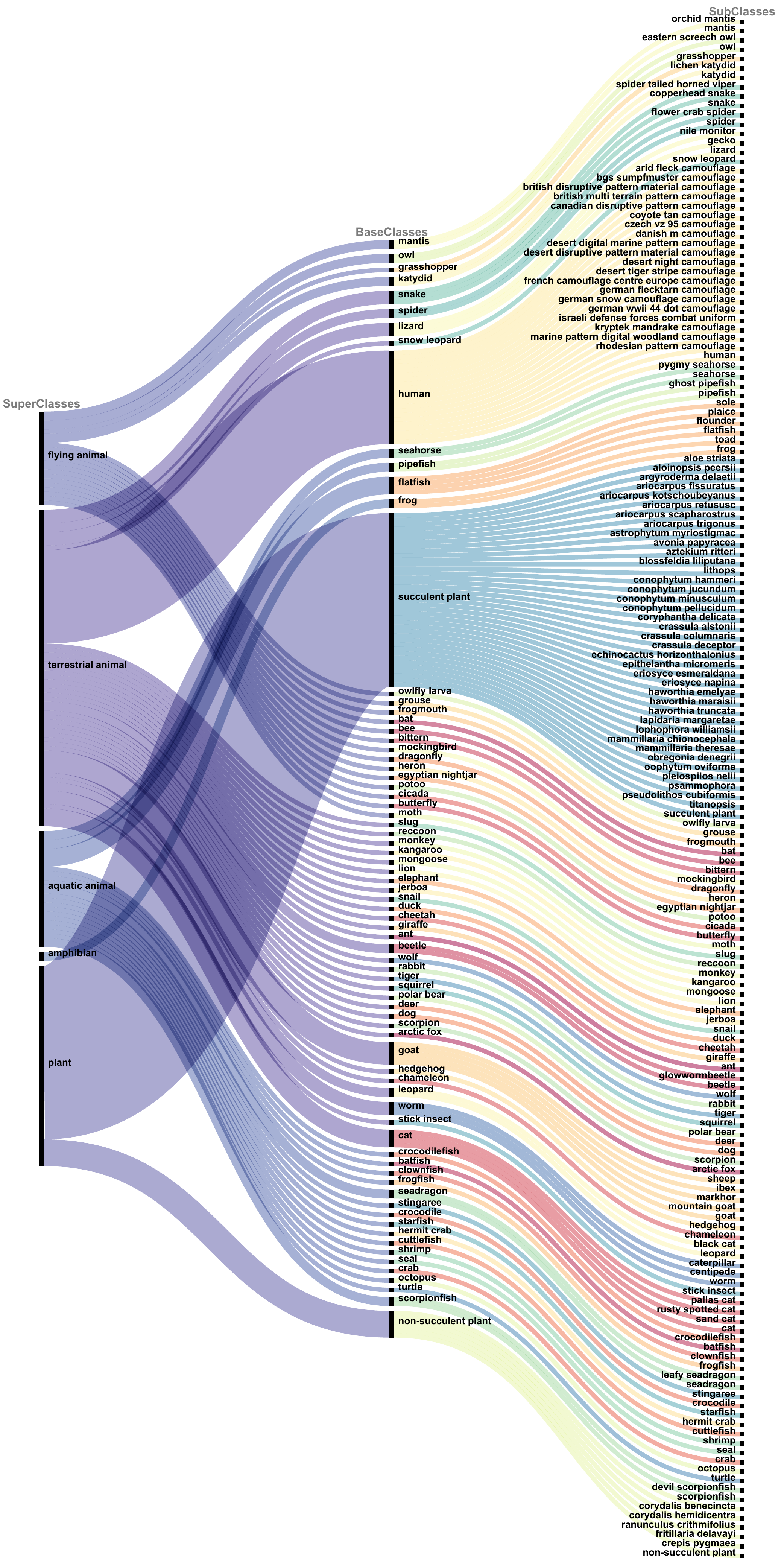}
  \caption{Hierarchy of sample classes contained in the proposed \mydataset.
    \textbf{Only the base classes are used in our experiments.}
    Sub-classes that do not meet the criteria are simply removed and are not displayed here.
  }
  \label{fig:class_alluvival}
\end{figure}

In~\cref{fig:class_alluvival}, we use the alluvial graph to show the class relationships at different levels, including super, base, and sub-classes.
The sub-classes shown here include class names with clearer meanings preserved from the original data and class names after initial manual correction.
The base classes represent the class names obtained after careful manual filtering and merging, which was used in all the experiments in this paper.
The super classes generalize the base classes from a broader perspective.

\section{Limitations and Future Works}
\label{sec:limitation}

\subsection{Class Embedding Setting}
\label{sec:class_embedding_setting}

In our proposed algorithm, the class embedding setting follows the existing OVSIS methods.
Although our iteration process does not impose much computational burden due to our caching mechanism, there is still a computational complexity associated with the number of classes.
This is not ideal for practical applications in open vocabulary scenarios.
More flexible and efficient class embedding designs are still worth exploring.

\subsection{CLIP-based Architecture}
\label{sec:clip_based_architecture}

The transfer application of the CLIP in downstream dense prediction tasks is limited by its pre-training form and the camouflage scenes that we focus on may be more affected.
Some work~\cite{OVSeg-OVSeg,OVSeg-SAN,OVSeg-CATSeg} attempts to further finetune CLIP, but the finetuning strategy still needs to be designed more carefully due to the potential disruption to the CLIP's open vocabulary ability.
And it also suggests that there is room for further improvement.
Besides, as mentioned in the main text, the current ideal performance of the CLIP-based architecture is still far from the limit, which suggests that future breakthroughs in this field may require more powerful paradigms.

\subsection{Data Scale}
\label{sec:data_scale}

Although the number of finely labeled samples in our proposed dataset is over ten thousand, the data scale is still smaller than existing large datasets such as COCO-Stuff~\cite{COCO-Stuff} and ADE20K~\cite{ADE20K}.
So there is still a need to collect more data, especially for classes with fewer samples.
We can resort to rich web images, which also may lead to significant manual labeling costs.
In addition, the data synthesis technique has been demonstrated in some recent work~\cite{FreeMask,SegGen} to greatly facilitate the performance of semantic image segmentation tasks.
This technique based on object masks and textual descriptions in existing datasets, may bring some new insights.
But this also needs to address the ensuing interference problem.

\subsection{Class Scale}
\label{sec:class_scale}

Open-vocabulary segmentation as a hot topic, currently focuses on how to use the open-vocabulary capability of VLMs to segment objects with unseen classes.
More data classes will indeed facilitate the development of OVCOS.
However, the imperceptibility of camouflaged objects poses a great challenge for further expansion.
This is the focus of our future work.

\subsection{Prompt}
\label{sec:prompt}

The importance of prompt engineering for visual language modeling can be reflected in the existing literature~\cite{CLIP,OVOD-ViLD,OVSeg-SimSeg,OVSeg-OVSeg,OVSeg-FCCLIP,OVSeg-ODISE,OVSeg-CATSeg,OVSeg-SAN} and the experiments in this paper.
The prompt forms used in this paper rely heavily on manual design, which is still limited by the knowledge of the prompter.
More automated prompt-generation strategies may be needed in the future, which deserve more attention and exploration.

\section{Dataset Copyright}
\label{sec:dataset_copyright}

We have investigated the copyright information of these data sources, and they are currently now widely used in the CSU field~\cite{DeepCSU}, and available for non-commercial academic use.
Much of existing work~\cite{ScribbleSupervisedSS,ScribbleSupervisedLiDARSS} provides only new annotations and an index of the original data rather than the data itself.
And users can download original data from the sources.
Following these existing community practices, for the proposed \mydataset dataset, we list the data links provided by the original authors in the documentation.
We also provide the class annotations we created and a detailed description of the way the data is organized in \mydataset.
In addition, we thank the contributors of the relevant datasets in the our acknowledgements.

\bibliographystyle{splncs04}
\bibliography{bib_main}
\end{document}

%% file: preamble.tex
\usepackage[dvipsnames]{xcolor}
\usepackage{booktabs}       %
\usepackage{multirow}
\usepackage{enumitem} %
\usepackage{colortbl} %

\definecolor{tabtitle}{gray}{.8}
\definecolor{ours}{gray}{.95}
\definecolor{ggray}{RGB}{127,127,127}
\definecolor{mygray}{RGB}{130,130,130}
\definecolor{mylred}{RGB}{250,130,130}
\definecolor{mylgreen}{RGB}{130,250,130}
\definecolor{mylblue}{RGB}{130,130,250}
\definecolor{reda}{RGB}{202,0,0}
\definecolor{redb}{RGB}{217,148,143}
\definecolor{myyellow}{RGB}{190,144,0}
\definecolor{mygreen}{RGB}{0,136,51}
\definecolor{myblue}{RGB}{0,102,204}

\definecolor{suppcolor}{RGB}{234,147,147}
\definecolor{truecls}{RGB}{0,255,0}
\definecolor{falsecls}{RGB}{0,255,255}

\usepackage{capt-of} %
\usepackage{pifont}
\newcommand{\none}{—}

\newcommand{\first}[1]{\textcolor{reda}{\textbf{#1}}}
\newcommand{\second}[1]{\textcolor{mygreen}{\textbf{#1}}}
\newcommand{\third}[1]{\textcolor{myblue}{\textbf{#1}}}

\newcommand{\vanilla}[1]{\textcolor{mygray}{#1}}

\newcommand{\parhead}[1]{\noindent\textbf{#1}}

\newcommand{\mytask}{{OVCOS}\xspace}
\newcommand{\mydataset}{{OVCamo}\xspace}
\newcommand{\mynetcore}{{semantic guidance and structure enhancement}\xspace}
\newcommand{\mymethod}{{OVCoser}\xspace}
\newcommand{\myprompt}{{CamoPrompts}\xspace}

\usepackage{bm}
\newcommand{\fontmod}[1]{\mathbf{#1}}
\newcommand{\fontcomp}[1]{\bm{#1}}

\newcommand{\fontset}[1]{\mathcal{#1}}

\usepackage{wrapfig}

\newenvironment{authorfoot}
  {}
  {}

%% file: table/object_attribute.tex
\begin{tabular}{l|>{\raggedright\arraybackslash}p{20em}}
  \toprule[2pt]
  \rowcolor{tabtitle}
  Attribute               & Description                                                                                                                        \\
  \midrule[1pt]
  Object Concentration    & Object pixel concentration, which calculates the area ratio between the object region and its minimum rotatable bounding box.      \\
  \midrule[0.5pt]
  Average Color Ratio     & Color ratio of the object to the background, which calculates the average of the three color channels in their respective regions. \\
  \midrule[0.5pt]
  Object-Image Area Ratio & Area ratio of the object relative to the image.                                                                                    \\
  \midrule[0.5pt]
  Number of Object Parts  & Number of separate areas in the image that belong to the object.                                                                   \\
  \midrule[0.5pt]
  Normalized Centroid     & Object centroid coordinates, which are normalized using the image shape.                                                           \\
  \bottomrule[2pt]
\end{tabular}

%% file: table/cmpsota_simple.tex
\begin{tabular}{c|c|c|c|*{6}{c}}
  \toprule[2pt]
  \rowcolor{tabtitle}
  \textbf{Model}                     & \textbf{VLM}                    & \textbf{Feature Backbone}                  & \textbf{Text Prompt}                  & cS$_{m}$ $\uparrow$ & cF$^{\omega}_{\beta}$ $\uparrow$ & cMAE $\downarrow$ & cF$_{\beta}$ $\uparrow$ & cE$_{m}$ $\uparrow$ & cIoU $\uparrow$ \\
  \midrule[1pt]
  \multicolumn{10}{l}{\textit{Test on \mydataset with the weight trained on COCO.}}                                                                                                                                                                                                                        \\
  \midrule[0.5pt]
  SimSeg$^{21}$~\cite{OVSeg-SimSeg}  & CLIP-ViT-B/16~\cite{CLIP}       & ResNet-101~\cite{ResNet}                   & Learnable~\cite{LearnablePrompt-CoOp} & 0.128               & 0.105                            & 0.838             & 0.112                   & 0.143               & 0.094           \\
  OVSeg$^{22}$~\cite{OVSeg-OVSeg}    & CLIP-ViT-L/14~\cite{CLIP}       & Swin-B~\cite{Swin}                         & \cite{OVOD-ViLD}                      & 0.341               & 0.306                            & 0.584             & 0.325                   & 0.384               & 0.273           \\
  ODISE$^{23}$~\cite{OVSeg-ODISE}    & CLIP-ViT-L/14~\cite{CLIP}       & StableDiffusionv1.3~\cite{StableDiffusion} & \cite{OVSeg-OpenSeg}                  & 0.409               & 0.339                            & 0.500             & 0.341                   & 0.421               & 0.302           \\
  SAN$^{23}$~\cite{OVSeg-SAN}        & CLIP-ViT-L/14~\cite{CLIP}       & ViT Adapter                                & \cite{OVOD-ViLD}                      & \third{0.414}       & \third{0.343}                    & \third{0.489}     & \third{0.357}           & \third{0.456}       & \second{0.319}  \\
  CAT-Seg$^{23}$~\cite{OVSeg-CATSeg} & CLIP-ViT-L/14~\cite{CLIP}       & Swin-B~\cite{Swin}                         & \cite{CLIP}                           & \second{0.430}      & \second{0.344}                   & \second{0.448}    & \second{0.366}          & \second{0.459}      & \third{0.310}   \\
  FC-CLIP$^{23}$~\cite{OVSeg-FCCLIP} & CLIP-ConvNeXt-L~\cite{OpenCLIP} & \none                                      & \cite{OVOD-ViLD}                      & 0.374               & 0.306                            & 0.539             & 0.320                   & 0.409               & 0.285           \\
  \midrule[0.5pt]
  \multicolumn{10}{l}{\textit{Finetune on \mydataset with the weight trained on COCO.}}                                                                                                                                                                                                                    \\
  \midrule[0.5pt]
  SimSeg$^{21}$~\cite{OVSeg-SimSeg}  & CLIP-ViT-B/16~\cite{CLIP}       & ResNet-101~\cite{ResNet}                   & Learnable~\cite{LearnablePrompt-CoOp} & 0.098               & 0.071                            & 0.852             & 0.081                   & 0.128               & 0.066           \\
  OVSeg$^{22}$~\cite{OVSeg-OVSeg}    & CLIP-ViT-L/14~\cite{CLIP}       & Swin-B~\cite{Swin}                         & \cite{OVOD-ViLD}                      & 0.164               & 0.131                            & 0.763             & 0.147                   & 0.208               & 0.123           \\
  ODISE$^{23}$~\cite{OVSeg-ODISE}    & CLIP-ViT-L/14~\cite{CLIP}       & StableDiffusionv1.3~\cite{StableDiffusion} & \cite{OVSeg-OpenSeg}                  & 0.182               & 0.125                            & 0.691             & 0.219                   & 0.309               & 0.189           \\
  SAN$^{23}$~\cite{OVSeg-SAN}        & CLIP-ViT-L/14~\cite{CLIP}       & ViT Adapter                                & \cite{OVOD-ViLD}                      & 0.321               & 0.216                            & 0.550             & 0.236                   & 0.331               & 0.204           \\
  CAT-Seg$^{23}$~\cite{OVSeg-CATSeg} & CLIP-ViT-L/14~\cite{CLIP}       & Swin-B~\cite{Swin}                         & \cite{CLIP}                           & 0.185               & 0.094                            & 0.702             & 0.110                   & 0.185               & 0.088           \\
  FC-CLIP$^{23}$~\cite{OVSeg-FCCLIP} & CLIP-ConvNeXt-L~\cite{OpenCLIP} & \none                                      & \cite{OVOD-ViLD}                      & 0.124               & 0.074                            & 0.798             & 0.088                   & 0.162               & 0.072           \\
  \midrule[0.5pt]
  \multicolumn{10}{l}{\textit{Train on \mydataset.}}                                                                                                                                                                                                                                                       \\
  \midrule[0.5pt]
  SimSeg$^{21}$~\cite{OVSeg-SimSeg}  & CLIP-ViT-B/16~\cite{CLIP}       & ResNet-101~\cite{ResNet}                   & Learnable~\cite{LearnablePrompt-CoOp} & 0.053               & 0.049                            & 0.921             & 0.056                   & 0.098               & 0.047           \\
  OVSeg$^{22}$~\cite{OVSeg-OVSeg}    & CLIP-ViT-L/14~\cite{CLIP}       & Swin-B~\cite{Swin}                         & \cite{OVOD-ViLD}                      & 0.024               & 0.046                            & 0.954             & 0.056                   & 0.130               & 0.046           \\
  ODISE$^{23}$~\cite{OVSeg-ODISE}    & CLIP-ViT-L/14~\cite{CLIP}       & StableDiffusionv1.3~\cite{StableDiffusion} & \cite{OVSeg-OpenSeg}                  & 0.187               & 0.119                            & 0.700             & 0.211                   & 0.298               & 0.167           \\
  SAN$^{23}$~\cite{OVSeg-SAN}        & CLIP-ViT-L/14~\cite{CLIP}       & ViT Adapter                                & \cite{OVOD-ViLD}                      & 0.275               & 0.202                            & 0.612             & 0.220                   & 0.318               & 0.189           \\
  CAT-Seg$^{23}$~\cite{OVSeg-CATSeg} & CLIP-ViT-L/14~\cite{CLIP}       & Swin-B~\cite{Swin}                         & \cite{CLIP}                           & 0.181               & 0.106                            & 0.719             & 0.123                   & 0.196               & 0.094           \\
  FC-CLIP$^{23}$~\cite{OVSeg-FCCLIP} & CLIP-ConvNeXt-L~\cite{OpenCLIP} & \none                                      & \cite{OVOD-ViLD}                      & 0.080               & 0.076                            & 0.872             & 0.090                   & 0.191               & 0.072           \\
  \rowcolor{ours}
  Ours                               & CLIP-ConvNeXt-L~\cite{OpenCLIP} & \none                                      & CamoPrompts                           & \first{0.579}       & \first{0.490}                    & \first{0.336}     & \first{0.520}           & \first{0.616}       & \first{0.443}   \\
  \bottomrule[2pt]
\end{tabular}

%% file: table/ablation.tex
\begin{tabular}{l|*{6}{c}|c}
  \toprule[2pt]
  \rowcolor{tabtitle}
  \textbf{Model}                                                       & cS$_{m}$ $\uparrow$ & cF$^{\omega}_{\beta}$ $\uparrow$ & cMAE $\downarrow$ & cF$_{\beta}$ $\uparrow$ & cE$_{m}$ $\uparrow$ & cIoU $\uparrow$ & $\mathbf{\Delta}$ \\
  \midrule[1pt]
  \multicolumn{8}{l}{\textit{Comparison of the proposed modules.}}                                                                                                                                                                        \\
  \midrule[0.5pt]
  Baseline                                                             & 0.517               & 0.408                            & 0.374             & 0.451                   & 0.549               & 0.359           & 0.0\%             \\
  +$\fontset{P}$                                                       & 0.543               & 0.435                            & 0.346             & 0.480                   & 0.581               & 0.383           & 6.3\%             \\
  +$\fontset{P},\fontcomp{C}$                                          & 0.550               & 0.453                            & 0.341             & 0.491                   & 0.597               & 0.397           & 9.1\%             \\
  +$\fontset{P},\fontcomp{C},\fontcomp{D}$                             & 0.565               & 0.473                            & 0.336             & 0.507                   & 0.606               & 0.422           & 12.6\%            \\
  +$\fontset{P},\fontcomp{C},\fontcomp{E}$                             & 0.567               & 0.481                            & 0.339             & 0.511                   & 0.607               & 0.432           & 13.5\%            \\
  +$\fontset{P},\fontcomp{C},\fontcomp{D},\fontcomp{E}$ (\ie, $T = 1$) & 0.570               & 0.488                            & 0.338             & 0.518                   & 0.610               & 0.436           & 14.5\%            \\
  \quad SE (\cref{fig:sea}) $\rightarrow$ Addition Fusion              & 0.552               & 0.457                            & 0.340             & 0.497                   & 0.599               & 0.402           & 9.9\%             \\
  \midrule[0.5pt]
  \multicolumn{8}{l}{\textit{Comparison of the proposed iterative refinement.}}                                                                                                                                                           \\
  \midrule[0.5pt]
  $T = 1$                                                              & 0.570               & 0.488                            & 0.338             & 0.518                   & 0.610               & 0.436           & 14.5\%            \\
  \rowcolor{ours}
  $T = 2$                                                              & \first{0.579}       & \first{0.490}                    & 0.336             & \first{0.520}           & \first{0.616}       & \first{0.443}   & \first{15.5\%}    \\
  \quad w/o $M_{cor}$ as in~\cref{fig:sga}                             & 0.575               & 0.487                            & 0.337             & 0.515                   & 0.611               & 0.441           & 14.8\%            \\
  \quad w/o $f_{obj}$ as in~\cref{fig:sga}                             & 0.571               & 0.476                            & 0.339             & 0.506                   & 0.608               & 0.434           & 13.4\%            \\
  $T = 3$                                                              & 0.576               & 0.484                            & \first{0.333}     & 0.514                   & 0.614               & 0.437           & 14.8\%            \\
  \midrule[0.5pt]
  \vanilla{$\lim\sup_{P_{seg} \rightarrow G_{seg}}$Perf.}              & \vanilla{0.703}     & \vanilla{0.703}                  & \vanilla{0.297}   & \vanilla{0.701}         & \vanilla{0.701}     & \vanilla{0.701} & \vanilla{51.2\%}  \\
  \bottomrule[2pt]
\end{tabular}

%% file: table/prompt_wmask1.tex
\begin{tabular}{c|l|c}
  \toprule[2pt]
  \rowcolor{tabtitle}
  \textbf{ID} & \textbf{Prompt Template}                         & $\mathcal{A}$ \\
  \midrule[1pt]
  0           & \texttt{"<class>"} w/o MAP based on ground truth & 0.538         \\
  \midrule[0.5pt]
  \multicolumn{2}{l}{\textit{Task-generic templates.}}                           \\
  \midrule[0.5pt]
  1           & \texttt{"<class>"}                               & 0.671         \\
  2.1         & \texttt{"The <class>."}                          & 0.648         \\
  2.2         & \texttt{"the <class>"}                           & 0.632         \\
  2.3         & \texttt{"A <class>."}                            & 0.677         \\
  2.4         & \texttt{"a <class>"}                             & 0.675         \\
  3.1         & \texttt{"A photo of a <class>."}                 & 0.684         \\
  3.2         & \texttt{"A photo of the <class>."}               & 0.691         \\
  3.3         & \texttt{"The photo of a <class>."}               & 0.672         \\
  3.4         & \texttt{"The photo of the <class>."}             & 0.680         \\
  3.5         & \texttt{"a photo of a <class>"}                  & 0.684         \\
  3.6         & \texttt{"a photo of the <class>"}                & 0.689         \\
  3.7         & \texttt{"the photo of a <class>"}                & 0.675         \\
  3.8         & \texttt{"the photo of the <class>"}              & 0.674         \\
  4           & templates from~\cite{CLIP}                       & 0.686         \\
  5           & templates from~\cite{OVOD-ViLD}                  & 0.682         \\
  \bottomrule[0.5pt]
\end{tabular}

%% file: table/prompt_wmask2.tex
\begin{tabular}{c|>{\raggedright\arraybackslash}p{18em}|c}
  \midrule[0.5pt]
  \multicolumn{2}{l}{\textit{Task-related templates.}}                                            \\
  \midrule[0.5pt]
  6.1  & \texttt{"A photo of the camouflaged <class>."}                                   & 0.681 \\
  6.2  & \texttt{"A photo of the concealed <class>."}                                     & 0.674 \\
  6.3  & 6.1, 6.2                                                                         & 0.685 \\
  6.4  & \texttt{"A photo of the <class> camouflaged in the background."}                 & 0.684 \\
  6.5  & \texttt{"A photo of the <class> concealed in the background."}                   & 0.694 \\
  6.6  & 6.4, 6.5                                                                         & 0.694 \\
  6.7  & \texttt{"A photo of the <class> camouflaged to blend in with its surroundings."} & 0.700 \\
  6.8  & \texttt{"A photo of the <class> concealed to blend in with its surroundings."}   & 0.697 \\
  6.9  & 6.7, 6.8                                                                         & 0.703 \\
  \rowcolor{ours}
  6.10 & 6.1 - 6.9 \ie,~\myprompt                                                         & 0.704 \\
  \bottomrule[2pt]
\end{tabular}

%% file: table/efficiency.tex
\begin{tabular}{c|c|c|c}
  \toprule[2pt]
  \rowcolor{tabtitle}
  \textbf{Model}                     & \textbf{Trainable Param.} & \textbf{Total Param.} & \textbf{FLOPs} \\
  \midrule[1pt]
  SimSeg$^{21}$~\cite{OVSeg-SimSeg}  & 61M (28.91\%)             & 211M                  & 1.9T         \\
  OVSeg$^{22}$~\cite{OVSeg-OVSeg}    & 531M (100.00\%)           & 531M                  & 8.0T         \\
  ODISE$^{23}$~\cite{OVSeg-ODISE}    & 28M (1.80\%)              & 1522M                 & 5.5T         \\
  SAN$^{23}$~\cite{OVSeg-SAN}        & 9M (2.06\%)               & 437M                  & 0.4T         \\
  CAT-Seg$^{23}$~\cite{OVSeg-CATSeg} & 104M (21.22\%)            & 490M                  & 0.3T         \\
  FC-CLIP$^{23}$~\cite{OVSeg-FCCLIP} & 20M (5.38\%)              & 372M                  & 0.8T         \\
  \rowcolor{ours}
  Ours                               & 7M (1.95\%)               & 359M                  & 0.2T         \\
  \bottomrule[2pt]
\end{tabular}